# A multi-algorithm approach for operational human resources workload balancing in a last mile urban delivery system

Luis M. Moreno-Saavedra[a], Silvia Jiménez-Fernández[a], José A. Portilla-Figueras[a], David Casillas-Pérez[b], Sancho Salcedo-Sanz[a]

[a]*Department of Signal Processing and Communications, Universidad de Alcalá, Alcalá de Henares, 28805, Madrid, Spain*
[b]*Department of Signal Processing and Communications, Universidad Rey Juan Carlos, Fuenlabrada, 28942, Madrid, Spain*

**Abstract**

Efficient workload assignment to the workforce is critical in last-mile package delivery systems. The explosive increase of e-commerce and last-mile package logistics after the COVID-19 pandemic has produced, among other issues, difficulties in balancing the daily workload of the workforce in many delivery zones. In this context, traditional methods of assigning package deliveries to workers based on geographical proximity can be inefficient and surely guide to an unbalanced workload distribution among delivery workers. In this paper, we look at the problem of operational human resources workload balancing in last-mile urban package delivery systems. The idea is to consider the effort workload to optimize the system, i.e., the optimization process is now focused on improving the delivery time, so that the workload balancing is complete among all the staff. This process should correct significant decompensations in workload among delivery workers in a given zone. Specifically, we propose a multi-algorithm approach to tackle this problem. The proposed approach takes as input a set of delivery points and a defined number of workers, and then assigns packages to workers, in such a way that it ensures that each worker completes a similar amount of work per day. The proposed algorithms use a combination of distance and workload considerations to optimize the allocation of packages to workers. In this sense, the distance between the delivery points and the location of each worker is also taken into account to minimize the travel time of the workers. The proposed multi-algorithm methodology includes different versions of k-means, evolutionary approaches, recursive assignments based on k-means initialization with different problem encodings, and a hybrid evolutionary ensemble algorithm. We have successfully illustrated the performance of the proposed approach in a real-world problem of human resource balancing in an urban last-mile package delivery workforce operating at Azuqueca de Henares, Guadalajara, Spain.

*Keywords:* Last mile package delivery, workload balancing, Evolutionary algorithms, Recursive algorithms, k-means
*PACS:* 0000, 1111
*2000 MSC:* 0000, 1111

## 1. Introduction

Last-mile package delivery refers to any logistic activity associated with packages (or express postal) delivery to private customer households in urban areas [1]. With the increase of e-commerce and delivery services in the last few years, highly pushed up by the COVID-19 pandemic, it is currently a hot topic in cities all over the world. Different problems related to this topic have been pointed out as key issues to make this activity more efficient and sustainable. Among them, the management of the unstoppable increase of the demand, the high cost associated with traditional home delivery, and different problems related to the workforce are some of the most important issues to be tackled in the years to come [2].

Many of these key problems related to last-mile delivery logistics can be solved by modern optimization techniques or Machine Learning (ML) approaches [3, 4]. Among them, very recent works have dealt with specific problems such as parcel locker location to improve last-mile delivery [5, 6, 7], vehicle routing problems for last-mile delivery using bikes [8], cars and drivers [9, 10], or multi-vehicle [11]. Other approaches have carried out research on different last-mile delivery models, including the use of drones [12], trucks and drones [13], cargo tunnels [14], same-day-delivery problems [15], last-mile green logistics [16], emergency logistics [17, 18], among others, and also human resources planning in last-mile delivery systems, which is fully related to the problem we are interested in this paper.

Despite that many researchers and practitioners are currently working on the complete automatization of last-mile delivery systems [19, 20], the reality is that from 2023, the large majority of the delivery is carried out by a human workforce. The COVID-19 pandemic changed the scene, and the importance of the workforce in different last-miles delivery models and systems was truly evident [21], from food delivery [22] to last-mile package and postal delivery systems [23]. In this context, different papers have dealt with problems related to the improvement of human resources workload and conditions in the last-mile delivery sector. Some of the most recent works discuss issues in crowdsourced delivery, i.e., to enable customers to be served efficiently and flexibly by occasional delivery worker. In [24], a logistic regression model is proposed to simulate the crowd agents' willingness to undertake a delivery, assuming that not all the proposed delivery tasks will necessarily be accepted. The model is completed with a novel compensation scheme, that determines reward values and a direct search algorithm for minimizing the expected total delivery cost. In [25], a theoretical acceptance model of crowdsourced delivery service is proposed. It includes factors such as trust, social influence, and loss of privacy in the model. The second big research area in the last years related to human resources in last-mile delivery problems is workforce balancing. The explosive increase in e-commerce and last-mile package logistics has led to difficulties in balancing the daily workload of the workforce, both in pick-up and delivery zones. In [26], an operational workload balancing problem in the context of order-picking is tackled. A mathematical model is proposed to obtain a new order-picking planning problem, together with an iterated local search algorithm to effectively set a fair operational workload balance. In [27], the effectiveness of different workload balancing approaches, such as the Rawlsian's approach, range or mean-based, among others, is evaluated also considering order-picking operations. Specifically, that work is focused on the context of balancing workload in cases of restricted time windows for retrieving customers' orders. In [28], a data clustering using a k-means algorithm is used, together with a Tabu Search approach to evaluating employees and jobs matching performance, in a human resource planning problem, directly applicable to last-mile delivery systems. In [29], the problem of workforce balancing in an urban delivery context is tackled. The work considers two types of workload: incentive workload, related to the delivery quantity (mainly affecting the courier's income), and effort workload, related to the delivery time (mainly affecting the courier's health). As fully discussed in [29], incentive workload has to be balanced over a long period (e.g., a week or a month), whereas effort workload has to be balanced over a short period (typically a shift or a day). That work formulates the workforce balancing problem by considering a multi-period workload, stochastic demand, and dynamic daily dispatching processes.

In this paper, we tackle a problem fully related to operational human resources in a last-mile urban package and express postal delivery. Specifically, we deal with a problem of operational human resources workload balancing, in terms of effort workload. The main objective of our proposal is to optimize the distribution of daily package deliveries among a pre-defined number of workers in a specific area. The objective is to minimize the variance of the workload distribution among the workers so that each worker works for approximately the same amount of time (optimization of the e"ort workload). The problem tackled is constrained by the number of workers available and, of course, by the location and number of delivery points in the study area. The solution should provide an optimal assignment of delivery points to workers, taking into account the distance among the delivery points, the workload of each delivery point, and the work area of each worker. The problem is a challenging NP-hard problem due to the combinatorial nature of the assignment problem, the dynamic nature of the package delivery industry, and the need to balance e!ciency and fairness in the distribution of work among workers. We propose a multi-algorithm approach for solving the problem, including k-means, evolutionary approaches, recursive assignments based on k-means initialization and di"erent problem encodings, and a hybrid evolutionary ensemble method. We have successfully illustrated the performance of the proposed approach in a real-world problem of human resource balancing in an urban last-mile package delivery workforce operating at Azuqueca de Henares, Guadalajara, Spain. We have compared the results obtained with those by alternative state-of-the-art algorithms, getting highly competitive results.

The contributions of this paper can be summarized as follows:

- We define and address an operational human resources workload balancing problem, a kind of clustering task with a specific work-time function, using a combination of distance and workload considerations. The final objective of the problem tackled consists of ensuring that each delivery worker works a similar amount of time in a last-mile urban delivery system.

- We propose a multi-algorithm approach based on hybridizing evolutionary computation techniques and recursive/iterative algorithms based on the k-means algorithm.

- We show that classical clustering algorithms, such as the k-means algorithm or Gaussian Mixture Models, among others, can be used to obtain solutions to the human resources workload balancing problem. However, the performance of these approaches is relatively poor, and hybridization with an evolutionary algorithm is needed to obtain near-optimal solutions to the problem.

- Experiments performed on real-world data have been analyzed to prove the applicability of the developed algorithms. The results obtained show a very good performance of the proposed multi-algorithm approach in di"erent real-world workload balancing problems in the context of last-mile urban delivery systems.

The remainder of the paper has been structured in the following way: the next section presents the problem definition, focused on obtaining an operational human resources workload balancing in terms of e"ort workload. Section 3 describes the proposed multi-method approach, with details of all the algorithms used and hybridized in this work. Section 4 presents the experimental part of the paper, where we discuss the results obtained by the proposed multi-method approach in a real-world problem of human resource balancing in an urban last-mile package delivery at Azuqueca de Henares, Guadalajara, Spain. Finally, Section 5 closes the paper with some ending conclusions and remarks on the research carried out.

## 2. Problem definition

As previously mentioned, we deal with a specific operational human resources workload balancing problem in a last-mile urban delivery system. The solution obtained by the proposed algorithms should provide an optimal daily assignment of delivery points to workers, taking into account the distance among the delivery points, the workload of each delivery point, and the work area of each worker (to promote the delivery persons' expertise in the area).

Figure 1 shows a simplified visual diagram of the operational human resources workload balancing tackled. At the top side of the figure, we represent the solution obtained by many standard optimization algorithms based only on distance. In this case, the workload balance is not achieved due to the distance-based metrics considerations in many traditional algorithms, such as clustering approaches. At the bottom of the diagram, it is possible to see a better approach that considers not only distance restrictions but also workload balance metrics or clusters' shape to get a balance between workers. We describe here two possible encodings to tackle the problem with algorithms, and then we will show the objective function proposed for this particular problem.

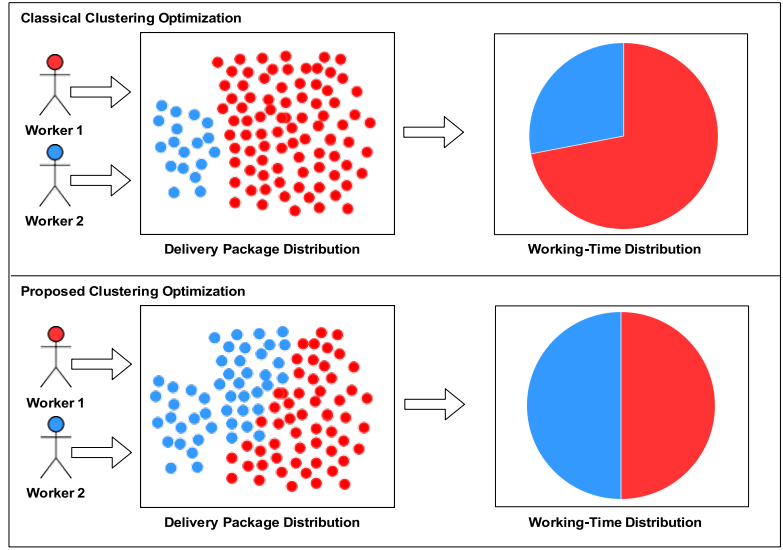

Figure 1: Diagram of operational human resources workload balancing.

*2.1. Problem encoding*

We have considered two di"erent possible encodings for the present problem: an *integer encoding* and a *circle encoding*. For both encodings, the proposed algorithms have been adapted with specific operators, as we will show later on.

The first encoding used to describe the problem in mathematical terms is the *integer encoding*. The individuals or solutions are vectors of size $N_p$, where $N_p$ is the total number of packages to be delivered:

$$\mathbf{x} = (x_1, x_2, \ldots, x_{N_p}), \tag{1}$$

where $x_1, x_2, \ldots, x_{N_p} \rightarrow [1, N_W]$, and $N_W$ stands for the number of workers in the system. Note that in this encoding, each element of the vector indicates the worker responsible for delivering that package.

Figure 2 shows a chart where the vector encoding can be seen at the top and a representation of the possible solution at the bottom. Each worker is represented with a di"erent color.

The second encoding considered in this paper is called *circle encoding*, i.e., circle zones are defined for each worker to work within them. Thus, in this encoding, each individual or solution consists of a vector of length $3 \cdot N_W$, where $N_W$ stands for the number of workers in the system. For each worker, the solution contains three values:

- $C_x$: x-coordinate of the position of the circle center, measured in meters when working with UTM coordinates.

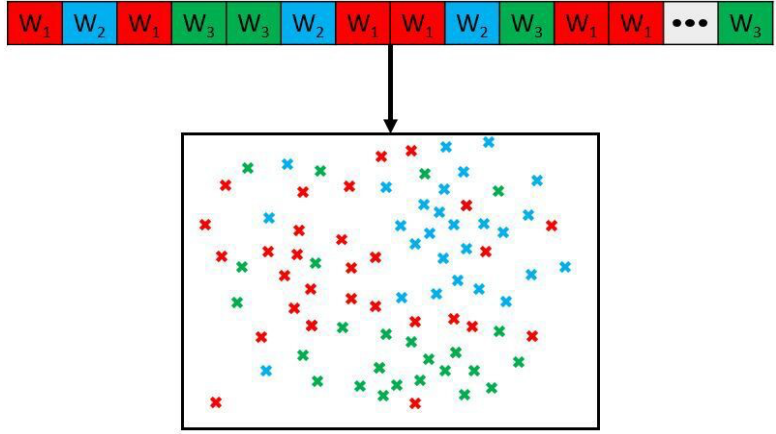

Figure 2: Integer Encoding Chart.

- $C_y$: y-coordinate of the position of the circle center, measured in meters when working with UTM coordinates.
- $R$: the radius of the circle, measured in meters.

Mathematically, this encoding is formally represented as follows:
$$\vec{\mathbf{x}} = \left[(C_{x_1}, C_{y_1}, R_1), (C_{x_2}, C_{y_2}, R_2), \ldots, (C_{x_{N_W}}, C_{y_{N_W}}, R_{N_W})\right], \tag{2}$$

where each triplet $(C_{x_i}, C_{y_i}, R_i)$ represents a circle zone defined for a given worker to deliver within.

Figure 3 shows a chart where the vector encoding can be seen at the top and a representation of the possible solution at the bottom. Each circle is represented with a di"erent color, and the delivery points inside each circle are assigned to that worker.

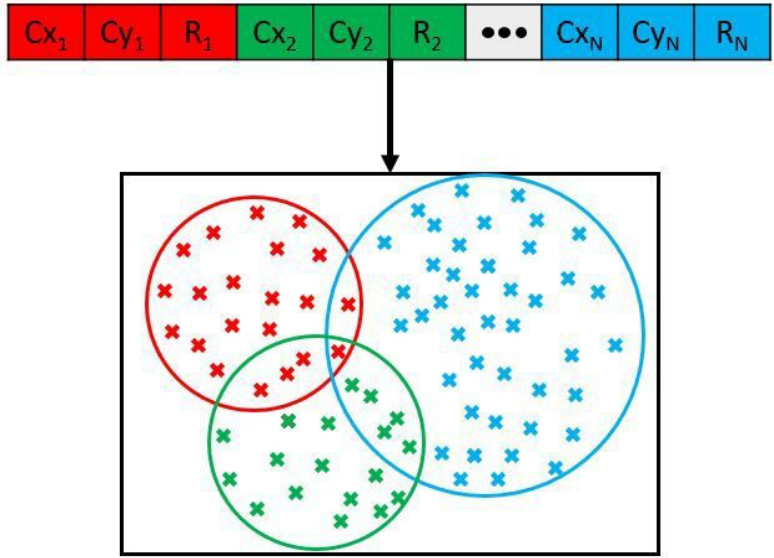

Figure 3: Circular Encoding Chart.

It is easy to see that the circle encoding ($\vec{\mathbf{x}}$) is a much more compact encoding than the integer encoding ($\mathbf{x}$): $3N_W$ versus $N_p$, as $N_p \gg 3N_W$ in the large majority of real-world cases. On the other hand, in the circle encoding, each worker is responsible for all the packages that fall within their corresponding circle, i.e., it provides an indirect assignment of the delivery packages to workers, whereas in the integer encoding, the allocation of packages to workers is direct in the encoding. Note that in this encoding, overlapping areas may appear in the solution. The treatment for the delivery packages in these overlapping areas consists of assigning the overlapped points to the last worker in the assignment round. With this assignment policy, we obtained better results in terms of fitness function value than using any other alternative policy, e.g., keeping the packages to the first worker or trying to distribute some packages to one worker and the rest delivery packages to other workers, among others.

It is not possible to know in advance which assignment is going to work better for a given task and algorithm (no-free lunch theorem [30, 31]), so we have analyzed the behavior of the proposed approaches with both encodings and have compared them in the experimental section of the paper.

*2.2. Objective function*

We define the objective function in this problem as the total working time ($t_W$) of each worker $j$, $t_W^j$. We want to minimize the difference in working time among the different workers so that they all work approximately the same amount of time. Several time inputs are taken into account to calculate the working time of each worker. The time inputs are as follows:

- One-Way Time ($t_{OW}$): It is the time it takes for the worker to go from the delivery center (where they pick up the packages / postal) to the first point of the delivery route.

- Return Time ($t_{Ret}$): It is defined as the time it takes for the worker to go from the last delivery point to the delivery center.

- Internal Handling Time ($t_{Int}$): It is the time it takes for the worker to prepare each order before starting the delivery route.

- External Handling Time ($t_{Ext}$): It is the time that it takes for the worker to deliver each delivery package. Note that this time includes waiting for the customer to open the door or requesting identification, among others.

- Travel Time ($t_{Tra}$): This time includes the travel time among all delivery points. We model it as the well-known Travelling Salesman Problem [32] (TSP).

Equation (3) shows how to calculate the total worker time for worker $j$ ($t_W^j$), with all times calculated in seconds:

$$t_W^j = t_{Int}^j + t_{OW}^j + t_{Tra}^j + t_{Ext}^j + t_{Ret}^j, \qquad 1 \leq j \leq N_W \tag{3}$$

where $N_W$ is the total number of workers in the system.

$$t_{Int} = \sum_{i=1}^{N_{Pj}} t_i^{In}, \tag{4}$$

$$t_{Ext} = \sum_{i=1}^{N_{Pj}} t_i^{Ex}, \tag{5}$$

where $N_{Pj}$ is the number of packages of each worker, $t_i^{In}$ is the internal handling time for each package and $t_i^{Ex}$ is the external handling time for each package. To calculate $t_{OW}$, $t_{Ret}$ and $t_{Tra}$, we use the distance traveled and the speed of each worker, defined as:

$$t = \frac{D}{s} \tag{6}$$

where *D* stands for the traveled distance and *s* stands for the worker speed to complete the distance. In this way, workers can be modeled in terms of the average worker speed they can achieve, and simulations can be run for workers on foot or workers driving different types of vehicles. Figure 4 shows an example of how the total worker time can be differently distributed for each worker depending on many factors, e.g., the distance between the work zone and the delivery center, the number of packages to be delivered or the spatial distribution of these packages.

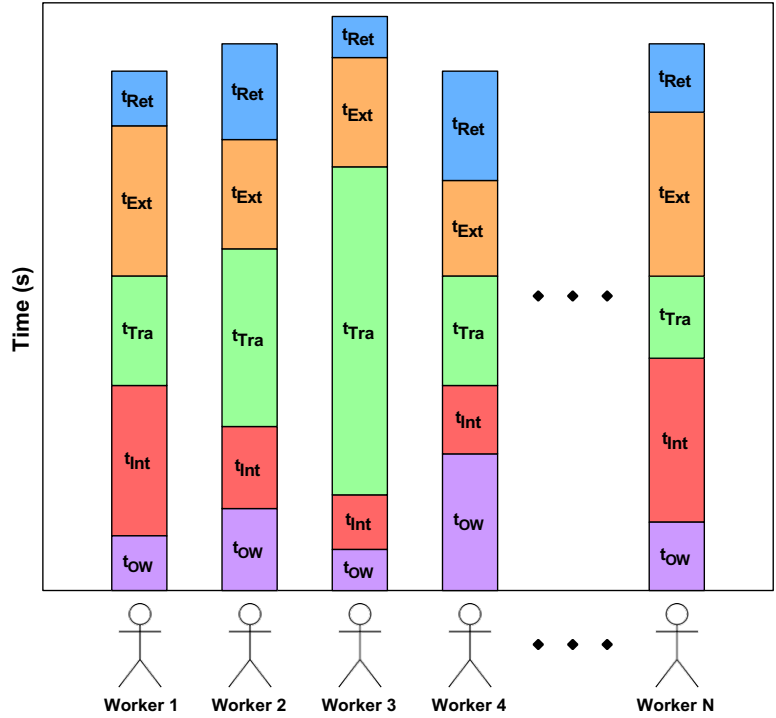

Figure 4: Example of the total time distribution for each worker of the delivery system. Note that colors are associated with specific times in the objective function ($t_{OW}$ (purple), $t_{Int}$ (red), $t_{Tra}$ (green), $t_{Ext}$ (orange) and $t_{Ret}$ (blue)). We will use this color convention in similar figures hereafter.

The final objective function used to compare each algorithm's performance is the difference between the worker *k* with the highest total working time and the worker *j* with the lowest total working time. Equation (7) shows the final objective function (note that it is the same objective function both for integer encoding (**x**) or circle encoding ($\vec{\mathbf{x}}$)):

$$g(\mathbf{x}) = \max(t_W^k) - \min(t_W^j), \quad 1 \leq k, j \leq N_W \qquad (7)$$

## 3. Proposed algorithmic approaches

In this section, we describe the different methods and algorithms proposed to solve the human resources workload balancing problem in a last-mile urban delivery system. In all cases, we consider that the number of delivery points is larger than the number of workers.

*3.1. Classic Clustering Algorithms*

In this section, we describe some Classic Clustering Algorithms that can solve the human resources workload balance problem at hand. The algorithms used are distribution-based clustering approaches because it is possible to specify the number of clusters before the algorithm starts. Specifying the number of clusters is necessary to determine the number of work zones, which is mandatory to be the same as the number of workers available. Algorithms such as k-Means [33], Gaussian Mixture Model [34], Balance Iterative Reducing and Clustering using

Hierarchies (BIRCH) [35], Agglomerative Hierarchy [36] and Spectral Clustering [37] have been implemented.

The k-means clustering algorithm is one of the most popular clustering algorithms nowadays. The algorithm works by iteratively assigning each data point to the cluster centroid with the lowest distance and recalculating the centroids as the average location of all the clustered points in the cluster. At the beginning of the algorithm, the centroids of the clusters are randomly assigned. The algorithm stops when the points assigned to each cluster do not change from one iteration to the next. The advantages of the k-means algorithm are its simplicity and e!ciency on large datasets. The disadvantages of this algorithm are its sensitivity to initial cluster centroids, which may lead to convergence problems and sub-optimal solutions, its scarce ability to handle di"erent types of data, and its assumption of spherical clusters due to the use of the distance to assign each point to a cluster.

Gaussian Mixture Models (GMMs) are probabilistic methods used for clustering and density estimation. GMMs assume that the data belong to a mixture of several Gaussian distributions, representing the searched clusters. Each cluster is defined by its mean, covariance matrix, and weight, which represents the probability that a point belongs to that particular cluster. The algorithm iteratively modifies these parameters using the Expectation-Maximization (EM) algorithm, which estimates the likelihood that each data point belongs to a particular cluster. One of the main advantages of GMMs is their ability to form non-spherical clusters, unlike the k-means algorithm. However, the computational cost of GMMs can be higher than that of k-means, especially for large numbers of clusters.

Balance Iterative Reducing and Clustering using Hierarchies (BIRCH) is a data clustering algorithm designed to work with large datasets, using a memory-e!cient approach by summarizing data points as Clustering Features (CFs) consisting of three components, i.e., a prototype point, a counter for the number of points represented, and a squared radius. The algorithm recursively builds a Clustering Feature Tree (CF Tree) to organize the data hierarchically. As more data points are encountered, BIRCH dynamically adjusts the branching and clustering parameters, changing the hierarchical structures of the clusters. The main advantages of BIRCH are related to computational time and memory usage for large datasets, as BIRCH can process the datasets in a single pass, reducing the need for multiple iterations. However, BIRCH may be less e!cient with irregularly shaped clusters or when the number of clusters is fixed.

Agglomerative Hierarchical Clustering is an algorithm that groups data into clusters by successively merging the points based on the distance. At the start, each data point is considered a single cluster and iteratively merges the closest clusters until only one cluster remains or until the specified number of clusters is reached. The result can be represented as a hierarchical tree-like structure called a dendrogram, which allows exploration of di"erent levels of granularity. The main advantages of Agglomerative Hierarchical Clustering are its ability to handle di"erent types of data and its intuitive representation of cluster relationships. The disadvantage of this algorithm is the high computational cost for significantly large datasets.

Spectral Clustering is an algorithm that uses the spectral properties of the data to cluster it. Spectral Clustering employs spectral decomposition to reduce the dimensionality space of the data, where clustering is performed using traditional techniques or algorithms such as k-means. The advantage of this algorithm is its ability to form complex cluster structures, such as non-convex or irregularly shaped clusters. However, its computational cost is higher than other algorithms like k-means for extremely large datasets.

*3.2. Evolutionary algorithm with integer encoding*

The first algorithm we propose is an Evolutionary Algorithm with Integer Encoding (EA-IE) [38, 39]. The EA is a well-known meta-heuristic optimization technique based on Darwin's principles of natural selection and genetic reproduction. Darwin's hypothesis of the principle of selection favors the perpetuation of the fittest individuals adapted to a given environment over generations. From an algorithmic point of view, an EA is an iterative optimization algorithm that uses this principle to search for global solutions. Figure 5 shows the workflow of the proposed EA with integer encoding.

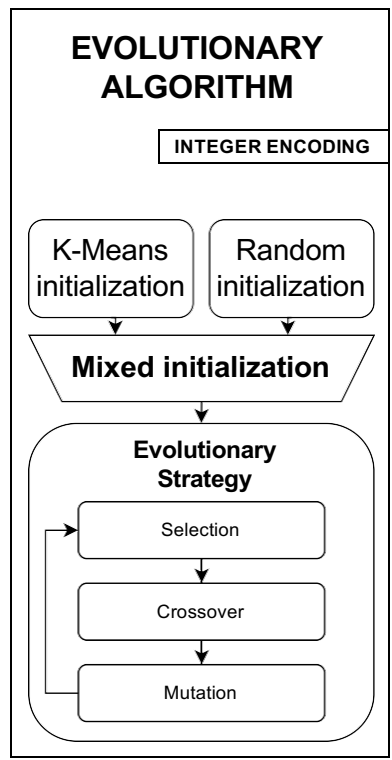

Figure 5: Workflow of the Evolutionary Algorithm with Integer Encoding.

As shown in Figure 5, the EA-IE algorithm starts from a set of initial solutions, encoded according to the integer encoding described above, and ordered according to their associated costs. Notice that half of the solutions are randomly initialized, and the rest are initialized using the k-means algorithm [33, 40]. The best solutions in the population, i.e., those with the lowest cost when the cost function is minimized, as in this case, are more likely to maintain their characteristics in the next iteration, as they are more likely to be selected to be reproduced, in a process known as *Crossover Operation*, which generates new solutions. In the crossover operation, the characteristics of the two parent individuals are mixed up to produce new individuals for the population. In addition, each time such individuals are generated, there is the possibility that they may undergo random alterations or mutations in the EA, in a process known as *Mutation Operation*. It consists of taking one or more characteristics of the individual and modifying the value by another random one. This *Mutation Operation* allows the EA to escape from local minima. New costs are assigned to the new individuals and compared with the original individuals in the population. The individuals with the worst costs in the population are discarded (*Selection operator*). This process is repeated until the maximum number of iterations is reached or until convergence. The Algorithm 1 also describes the EA workflow in detail.

The specific operators implemented in the EA with integer encoding are the following:

- Selection Operation: Once an iteration of the EA is finished, the selection operation takes place. In this operation, only the best adapted to the environment individuals of the population survive, i.e., the solutions with better objective function values.

- Crossover Operation: Once the selection operation has taken place, the next operation is

**Algorithm 1:** Workflow of the EA-IE algorithm.
    **Input:** Population $P$, initial population $P_0$, population size $N$, number of generations $G$
    **Output:** Best solution found $\mathbf{x_{best}}$
1   $t \leftrightarrow 0$;
2   **while** $t < G$ **do**
3      $P_t \leftrightarrow$ Selection individuals from $P_{t\uparrow 1}$;
4      $P_t \leftrightarrow$ Crossover operation between individuals in $P_t$;
5      $P_t \leftrightarrow$ Mutation operation of some individuals in $P_t$;
6      $\mathbf{x_{best}} \leftrightarrow$ Best individual in $P_t$;
7      $t \leftrightarrow t + 1$;
8   **end**
9   **return** $\mathbf{x_{best}}$;

the crossover operation. A multi-point crossover is implemented in our case. The multi-point crossover consists of generating a new individual ($\mathbf{y}$) by randomly selecting some genes from one individual ($\mathbf{x_1}$) and the rest of the genes from the other individual ($\mathbf{x_2}$). Figure 6 shows an example of a multi-point crossover operation.

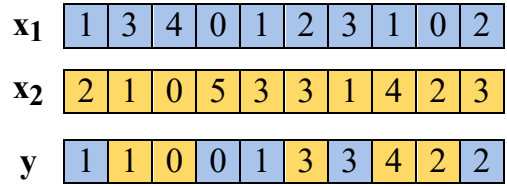

Figure 6: EA-IE crossover operation. $\mathbf{y}$ is the resulting solution after the crossover.

- Mutation Operation: Each time a new individual is generated, there is a possibility that this individual will mutate. Mutation consists of replacing one of the gene values with a random value. Figure 7 shows an example of the mutation operation implemented in the EA.

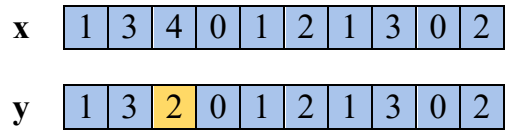

Figure 7: EA-IE mutation operator. $\mathbf{y}$ is the resulting solution after the mutation.

*3.3. Evolutionary algorithm with circle encoding*

The second algorithm implemented in this paper is an EA with circle encoding (EA-CE). Figure 8 shows the workflow of the proposed EA.

The main di"erence between the EA-CE and the EA-IE is the change in the type of encoding, which in turn has e"ects on the operators implemented. In the EA-CE, two di"erent types of crossovers and two di"erent types of mutations have been developed as adaptations to this particular encoding. The crossover and mutation operators developed for the EA-CE are as follows:

- External Crossover: This crossover is a multi-point crossover as explained above, but instead of changing single elements, whole circle regions represented by triplets ($Cx_i$, $Cy_i$, $R_i$)

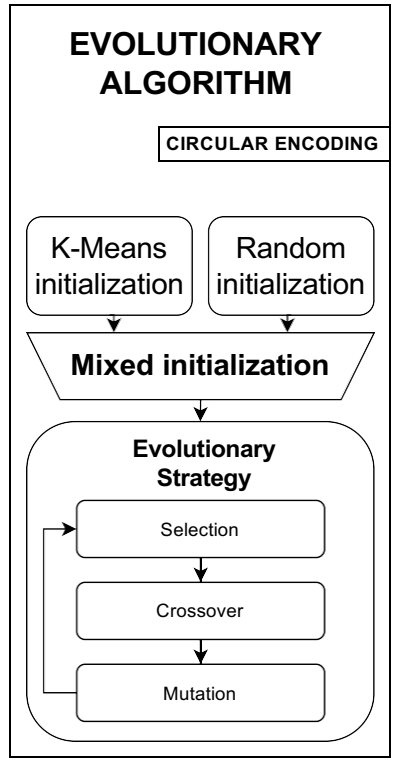

Figure 8: Workflow of the EA-CE algorithm.

are changed as a block. Figure 9 shows an example of External Crossover in EA with Circle Encoding.

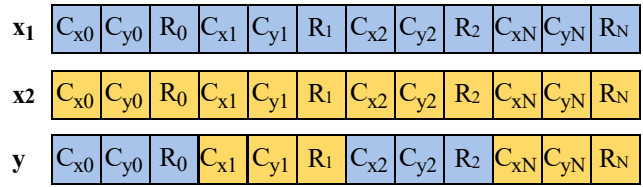

Figure 9: EA with Circle Encoding External Crossover Operator.

- Internal Crossover. This crossover is a multi-point crossover as explained above, but it treats the values ($Cx_i$, $Cy_i$, $R_i$) as if they were independent values. Figure 10 shows an example of internal crossover in the EA with circle encoding.

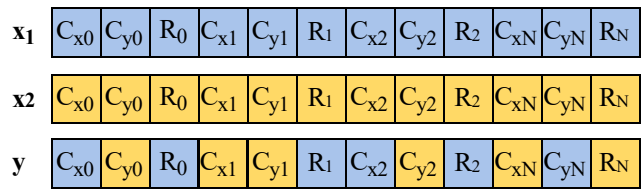

Figure 10: EA with circle encoding internal crossover operator.

- Smooth Mutation: This mutation has a medium-high probability of occurrence, and consists of modifying the values of the vector using a normal distribution. The new values are calculated as:

$$x^{\downarrow} = x + N(0, 1), \qquad (8)$$

Figure 11 shows an example of a smooth mutation in the EA with circle encoding:

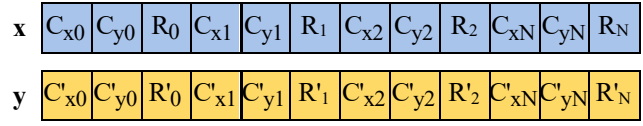

Figure 11: EA-CE smooth mutation operator.

- Hard Mutation. This mutation has a low probability of occurrence and consists of modifying a randomly-chosen triplet ($Cx_i$, $Cy_i$, $R_i$) with random values. Figure 12 shows an example of this hard mutation in the EA-CE.

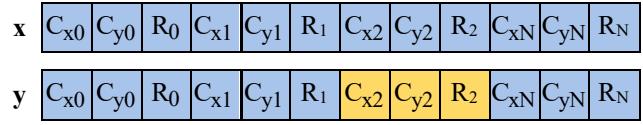

Figure 12: EA-CE hard mutation operator.

### 3.4. Recursive algorithm with k-means initialization and integer encoding

In this section, we discuss a recursive heuristic algorithm that uses k-means initialization and integer encoding, the RA-IE. Figure 13 shows the workflow of the RA-IE algorithm.

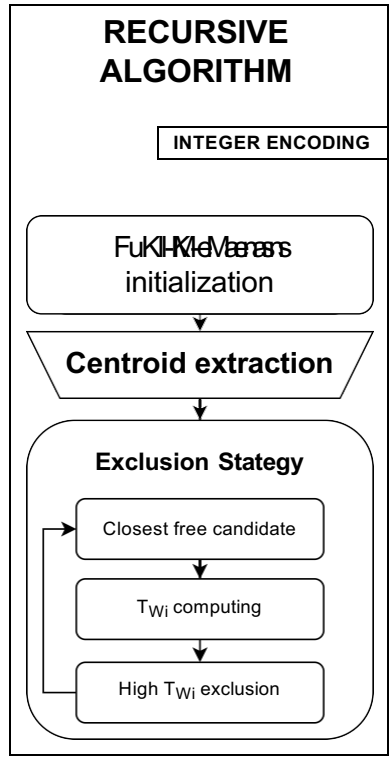

Figure 13: Workflow of the RA-IE algorithm.

After finishing the k-means algorithm (where $k$ is the number of centroids and is the same as the number of workers $N_W$), a clustering is obtained and used to initialize the RA-IE algorithm. Then, we select the centroids of the k-means algorithm-generated clusters and assign the nearest available point to each centroid. After one assignment round, we calculate each worker's work time using Equation (3) and the average work time before proceeding to the next assignment round. During the assignment round, workers whose work time exceeds the average work time will not receive any points. Only workers whose work time is below the average work time will receive delivery points to balance the workload. This process is repeated until all the delivery points have been allocated to the workers. Algorithm 2 shows the workflow of the RA-IE algorithm.

---

**Algorithm 2:** Workflow of the RA-IE algorithm.
---
**Input:** Initial delivery points $P_0$, number of workers $N_W$
**Output:** Best solution found $x_{best}$
1 Get centroids $\mu_1, \mu_2, \ldots, \mu_{N_W} \leftrightarrow$ k-means Algorithm;
2 Worker $W_i$ start zone $\leftrightarrow \mu_i$;
3 Worker's work time $t_{W_i} \leftrightarrow 0$;
4 **while** *Available Points in $P_0$* **do**
5     **for** *each centroid $\mu_i$* **do**
6         **if** $t_{W_i} < t_{W_{mean}}$ **then**
7             $\mu_i \leftrightarrow$ Nearest available point;
8             Calculate new $t_{W_i}$;
9         **end**
10     Calculate new $t_{W_{mean}}$;
11 **end**
12 **return** $x_{best}$;

---

*3.5. Recursive algorithm with k-means initialization and circle encoding*

In this section, we discuss another recursive algorithm that uses k-means as initialization and circle encoding (RA-CE). Figure 14 shows this approach's workflow.

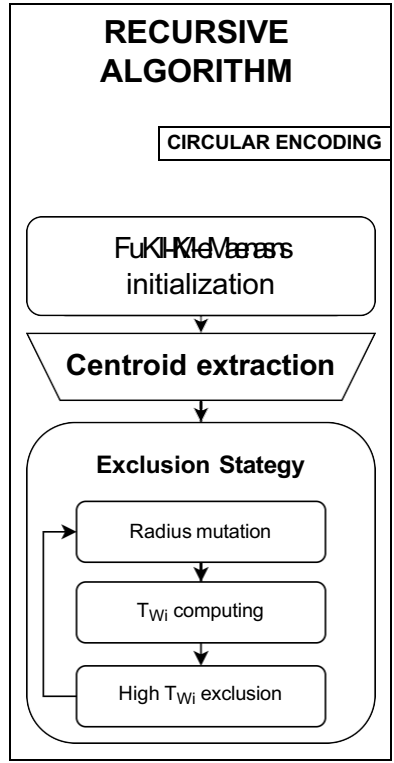

Figure 14: Workflow of the RA-CE algorithm.

The RA-CE is an algorithm that hybridizes the methodology followed in the RA-IE (see Section 3.4) and the operators used in the EA with circle encoding (see Section 3.3). Therefore, we create enclosed workspaces in which the assignment of the delivery points is limited to those that belong to the inside of the circumference. After finishing the k-means algorithm (where $k$ is the number of centroids and is the same as the number of workers $N_W$), a clustering is obtained and used to initialize the RA-CE algorithm. Then, we determine the centroids of the k-means algorithm-generated clusters. These centroids will correspond to the centers of the circles that will delimit the work zones. The radius of each circle is randomly selected. For each of the work zones, the time of each worker is calculated using Equation (3), taking into account that the distribution points are all those inside the circumference that delimits the work zone. Once all worker's working times have been calculated, the average working time is obtained in the same way as in the RA-IE. After an assignment round, the radius of the circles will vary according to their working time compared to the average time. If a worker's working time is less than the average time, the radius of the circumference is increased to cover more delivery points. Analogously, if a worker's working time is above the average time, the radius is decreased to include fewer delivery points, thus balancing all workers' working time. Di"erent tests have been run and a 3% radius increase/decrease has been set. Therefore, when the circle has to gather more points, the radius is multiplied by 1.03, while it is multiplied by 0.97 when fewer delivery points need to be considered. The algorithm ends when all delivery points are assigned. Algorithm 3 shows the workflow of the RA-CE algorithm.

---

**Algorithm 3:** Workflow of the recursive algorithm with k-means initialization and circle encoding.

**Input:** Initial Delivery Points $P_0$, Number of workers $N_W$
**Output:** Best solution found $\mathbf{x}_{best}$

1 Get centroids $\mu_1, \mu_2, \ldots, \mu_{N_W}$ ↔ k-means Algorithm;
2 Worker $W_i$ circle center ↔ $\mu_i$;
3 Radius of the circle $R_i$ ↔ Random value;
4 Calculate worker's work time $t_{W_i}$;
5 **while** *Available Points in $P_0$* **do**
6     Calculate new $t_{W_{mean}}$;
7     **for** *each centroid $\mu_i$* **do**
8         **if** $t_{W_i} < t_{W_{mean}}$ **then**
9             $R_i^{\downarrow}$ ↔ $1.03 \cdot R_i$;
10             Calculate new $t_{W_i}$;
11         **else if** $t_{W_i} > t_{W_{mean}}$ **then**
12             $R_i^{\downarrow}$ ↔ $0.97 \cdot R_i$;
13             Calculate new $t_{W_i}$;
14     **end**
15 **end**
16 **return** $\mathbf{x}_{best}$;

---

*3.6. Ensemble final EA approach*

The Ensemble Final EA approach consists of the hybridization between the Evolutionary Algorithm with Integer Encoding (EA-IE) (see Section 3.2) and the Recursive Algorithm with k-means Initialization and Integer Encoding (RA-IE) (see Section 3.4), to form the so-called RA-EA-IE algorithm. Note that the proposed EA shows good performance in terms of cost in this problem. However, random initialization of the algorithm leads to computational and convergence problems. To try to improve the algorithm's convergence and get better results, first, we apply the fast Recursive Algorithm described above, and we use its solutions and some variations of these solutions to seed the EA first population. This way, the EA should be able to start from better solutions and evolve them to reach better results than starting from a population of randomly generated individuals. As we will show later in the experimental part, this strategy is very e"ective and leads to better results of the EA approach in this problem. Figure 15 shows the workflow of this ensemble approach.

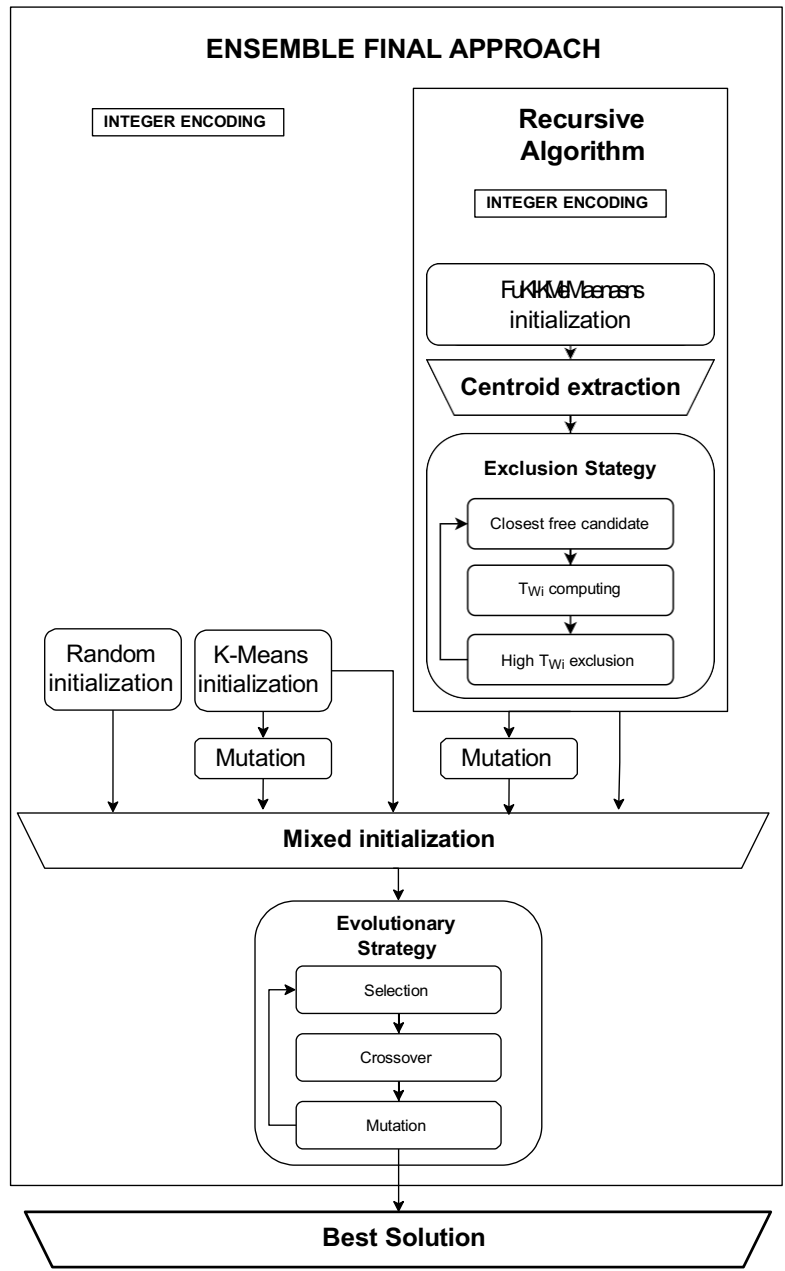

Figure 15: Workflow of the ensemble RA-EA-IE algorithm.

The first step in the RA-EA-IE algorithm is to run several times the RA-IE algorithm to obtain several individuals that present a good quality fitness function value. In the same way, the k-means algorithm is executed, and several individuals are obtained. Then, the EA-IE is started considering the following initial population:

- A percentage of the population is randomly generated using a uniform distribution.

- A percentage of the population is generated using k-means individuals.

- A percentage of the population is generated using k-means mutated individuals.

- A percentage of the population is generated using RA-IE solutions.

- A percentage of the population is generated using RA-IE mutated individuals.

Once the initial population is obtained, the EA proceeds as in the EA-IE (see Algorithm 1).

## 4. Experiments and Results

In this section, we present the experimental part of the paper. We first describe the data that has been used to carry out this study that correspond to real-world data from Azuqueca de Henares, Guadalajara, Spain. Then, we present and discuss the results obtained in two different experiments: in the first experiment, a comparison of the different methodologies explained in Section 3 is developed (considering days with different workload: low, medium, and high). In the second experiment, we show the results of the best-performing algorithm (RA-EA-IE) for 12 consecutive days. To finish this chapter, we present a section that discusses additional results. Furthermore, we show the experiments carried out and the results obtained about the Classical Clustering Algorithms explained in Section 3.1.

### 4.1. Data available and experiments description

Data available for this research correspond to packets and post-mail deliveries to homes and stores in Azuqueca de Henares, Guadalajara, Spain. Correos group (the former state post service in Spain, now converted into a reference logistics company) provides real-world data through the Research Chair Correos has at the University of Alcalá. These data have been averaged over a week and randomized to preserve privacy. Correos works and has offices in the entire Spanish territory, including big, medium, and small cities. In this case, Azuqueca de Henares is a medium city in Spain, located in the province of Guadalajara, belonging to the autonomous community of Castilla-La Mancha. Azuqueca de Henares is the second most important town of Guadalajara by population and socioeconomic development. The city has a population of 35182 inhabitants and an area of 19.68 $km^2$. It is located at the coordinates $40°\,33'\,53''$ N and $3°\,16'\,05''$ W. Figure 16 shows some geographical information about the dataset used in this problem. Figure 16a shows the location of Azuqueca de Henares on a map of Spain, and Figure 16b shows the sites of all the houses and stores that can be potential sites for receiving a package (in green color). The position of the delivery center (Correos central post office in the city) is shown in red color. We have chosen the case of Azuqueca de Henares because it is the paradigm of a medium town in Spain with one central post office. Bigger cities usually have more than one distribution office, so the delivery is split into different parts of the town, usually with fewer workers than in the case of medium cities such as Azuqueca. It is important to keep in mind that we have carried out the work-balancing of the delivery hours, not the total working hours: all the workers in Correos Azuqueca must spend part of their working day carrying out office work, and part as delivery workers. We have acted over this second part of their working hours, so the time spend for all of them as delivery workers is as balanced as possible.

We consider two different experiments to illustrate the performance of the algorithms developed for this problem of human resource workload balancing in last-mile urban delivery systems. In Experiment 1, we analyze the performance of the different proposed algorithms on three distinct days of package delivery in Azuqueca de Henares. Each day has a different number of workers and a different number of packages to be distributed. We select a day when the number of delivery packages to be distributed is low, a day when the number of delivery packages to be distributed is average, and a day where the number of delivery packages to be distributed is high compared with the average delivery package distribution in Azuqueca de Henares. Table 1 shows the delivery package distribution and the available workers to deliver

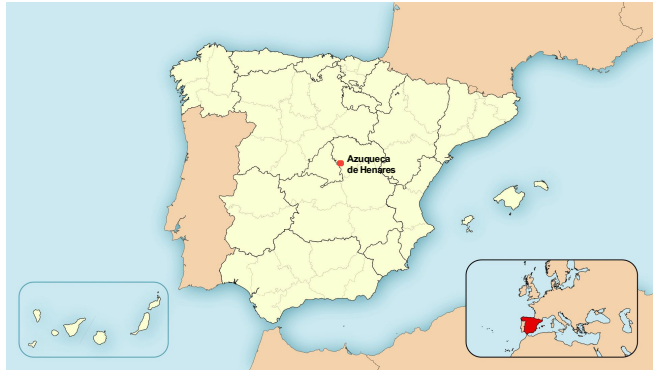

(a) Location of Azuqueca de Henares.

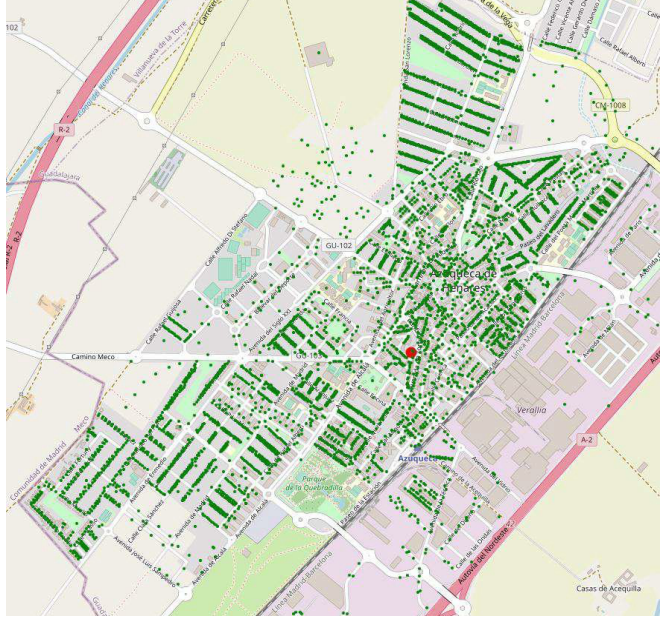

(b) Houses and stores in Azuqueca de Henares in green color and delivery center in red color.

Figure 16: Geographical information about the dataset.

the packages each day in consideration. Figure 17 graphically shows the delivery package distribution in Azuqueca de Henares for each day considered in Experiment 1. Green-colored points are homes and stores where packages must be delivered that day, while the red-colored point is the delivery center, where all the workers start their journey. It is possible to see that the density of the points to deliver on day 1 (Figure 17a) is lower than the density of points on day 2 (Figure 17b), and much lower than the density of points in day 3 (Figure 17c).

Table 1: Total number of packages and available workers for each day in Experiment 1. On day 1 the number of packages is low, on day 2 the number of packages is average, and on day 3 the number of packages is high.

|       | load    | packages | workers |
|-------|---------|----------|---------|
| Day 1 | low     | 240      | 12      |
| Day 2 | average | 392      | 12      |
| Day 3 | high    | 628      | 13      |

We consider a second experiment (Experiment 2), further to illustrate the behavior of the proposed multi-algorithm approach. In this case, the data correspond to 12 consecutive days of package delivery in Azuqueca de Henares. The second experiment objective is to verify the performance of the RA-EA-IE algorithm (see Section 3.6) against di"erent package delivery situations. Table 2 shows the packages to be delivered and the available workers for each day in Experiment 2. Figure 18 indicates the number of delivery packages (blue bars) and the number of workers (red line) for each simulation day. It is possible to see that the distribution of packages and the number of workers are di"erent among days so that the performance of the RA-EA-IE algorithm can be evaluated in distinct cases, e.g., day 1 has a high number of packages to be distributed by 13 workers and day 9 has a similar number of packages to be distributed with only 11 workers (real situations occurring in Correos o!ces due to di"erent circumstances, such as temporal workers absences due to sickness, among others). Another example of the variability occurs between days 2 and 3 and days 10 and 11. The number of packages between those days is similar, but the number of workers is distinct, with 12 workers in the first case and 13 workers in the second case.

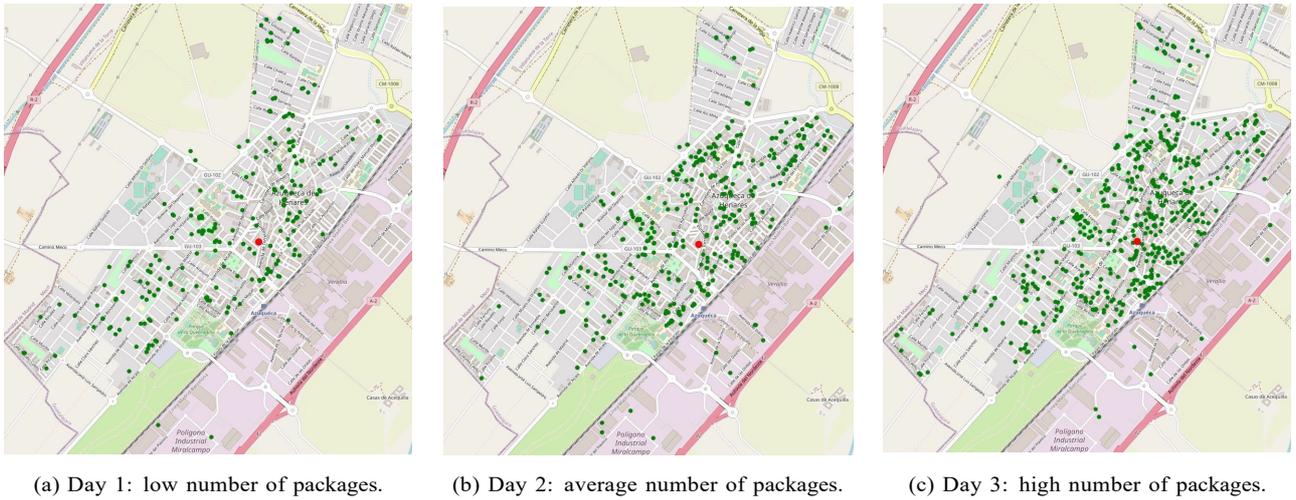

(a) Day 1: low number of packages.   (b) Day 2: average number of packages.   (c) Day 3: high number of packages.

Figure 17: Experiment 1 initial package distribution (green points) and delivery center (red point) in Azuqueca de Henares.

Table 2: Total number of packages and available workers for each day in Experiment 2.

|        | packages | workers |
|--------|----------|---------|
| **Day 1**  | 715 | 13 |
| **Day 2**  | 244 | 12 |
| **Day 3**  | 252 | 12 |
| **Day 4**  | 455 | 12 |
| **Day 5**  | 547 | 11 |
| **Day 6**  | 538 | 13 |
| **Day 7**  | 333 | 12 |
| **Day 8**  | 382 | 13 |
| **Day 9**  | 676 | 11 |
| **Day 10** | 285 | 13 |
| **Day 11** | 270 | 13 |
| **Day 12** | 645 | 13 |

In both experiments, the hyperparameters of the proposed algorithms, mainly in the case of EAs, have been experimentally set to obtain the best possible performance in terms of cost in the fitness function.

### 4.2. Results for Experiment 1

We describe here the results obtained in Experiment 1 at Azuqueca de Henares, aditionally the specific experiments carried out and the simulation parameters for each algorithm used.

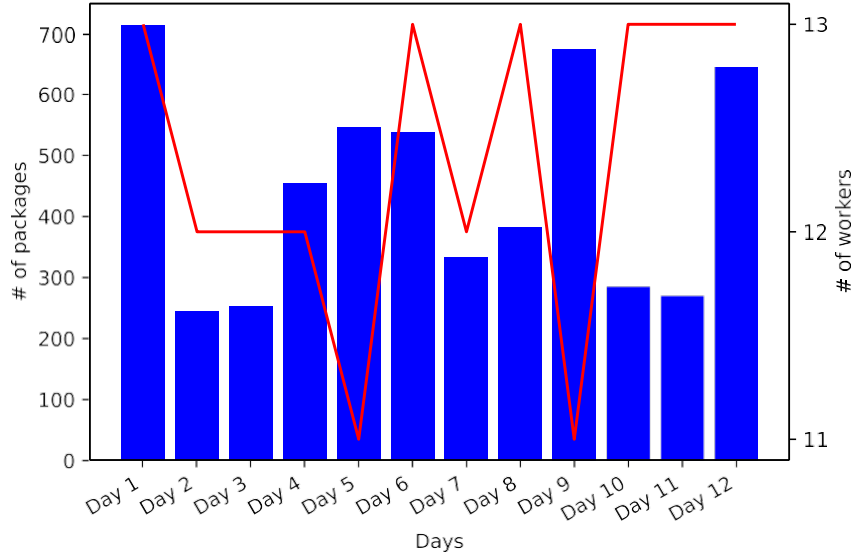

Figure 18: Distribution of the number of packages (blue bars) and the number of workers (red line) in Experiment 2.

To get statistically significant results, we ran 30 simulations of each algorithm and calculated various statistical parameters, e.g., minimum value, maximum value, mean value, and standard deviation for all the experiments. The algorithms used for the simulations are those described in Section 3, i.e., EA-IE, EA-CE, RA-IE, RA-CE, and RA-EA-IE. Notice that we have discarded the RA-EA-CE algorithm due to poor performance in previous experimental algorithm analysis. As already mentioned, each algorithm has been evaluated on three di"erent days, each representing a day with a distinct amount of packages to be delivered: a day with a low number of packages (type of day 1), a day with an average number of packages (type of day 2), and a day with a high number of packages to be delivered (type of day 3) (see Section 4.1).

Table 3 shows general information and other parameters used in each algorithm. The general information is about the number of total simulations for every algorithm and for each day, the maximum time for each simulation, and the algorithm used to compute the TSP, used to calculate the worker's route distance, all enabling the calculation of $t_{Tra}$. The parameters of each algorithm refer to the Internal Handling Time, $t_i^{In}$, for each package (used in Equation (4)), the External Handling Time, $t_i^{Ex}$, for each package (used in Equation (5)) and worker speed, $s_W$, (used in Equation (6) to calculate times related to distance).

Table 3: General information and fixed parameters for each algorithm in Experiment 1.

| | |
|---|---|
| **Number of simulations** | 30 |
| **Maximum Simulation Time** | 40 min. |
| **TSP Solver** | Local Search |
| $t_i^{In}$ | 57.64s |
| $t_i^{Ex}$ | 132.76s |
| $s_W$ | 5 km/h |

Table 4 shows the hyper-parameters of the Evolutionary Algorithms (EA-IE, EA-CE, RA-EA-IE). These parameters are specifically the number of individuals in the population, the number of generations or stop criteria to finish the algorithm, the mutation probability when a new individual is created, the crossover fraction for each parent individual to generate the

o"spring individual, and the survival fraction in the Selection operator at the start of a new generation.

Table 4: Hyper-parameters in Evolutionary Algorithms, i.e., EA-IE, EA-CE and RA-EA-IE.

| | |
|---|---|
| **Number of Individuals** | 40 |
| **Number of Generations** | 200 or max. time reached |
| **Mutation Prob.** | 0.05 |
| **Crossover Frac.** | 0.5 |
| **Survival Frac.** | 0.5 |

For each simulation, the outcome of each algorithm is every worker's total working time. These data are used to calculate the fitness function (see Equation (7)) for each simulation. Table 5 shows the results obtained for the first type of day (low delivery) for each considered algorithm, i.e., EA-IE, EA-CE, RA-IE, RA-CE, and RA-EA-IE. The results shown are the statistical data of the 30 simulations using the fitness function value. The RA-EA-IE algorithm obtained the best performance in terms of minimum fitness function value for all the simulations, compared to the other algorithms, the best in terms of mean value, and the best in terms of standard deviation. The second algorithm in terms of fitness function value is the EA-IE algorithm, followed by the RA-IE algorithm. The EA-IE algorithm gets better results than the RA-IE algorithm, but the advantage of the RA-IE algorithm is the total simulation time because the EA-IE algorithm needs a simulation time of 40 minutes, while the RA-IE algorithm spends only a few seconds to get the results. The last algorithms in terms of fitness function value performance are the algorithms that use circle encoding. It is possible to see that these algorithms, i.e., the RA-CE algorithm and the EA-CE algorithm, have, in general, a lower performance compared to the algorithms that use integer encoding. Thus, circle encoding is not appropriate for this particular problem or in that geographical distribution of the delivery packages.

Table 5: Experiment 1 fitness function statistical values for each algorithm considered for a day with a low number of packets to be distributed (type of day 1).

| | **Min.**(s) | **Max.**(s) | **Mean**(s) | **Std.**(s) |
|---|---|---|---|---|
| **EA-IE** | 308.49 | 1713.30 | 642.53 | 297.59 |
| **EA-CE** | 4013.24 | 6125.81 | 5201.63 | 514.21 |
| **RA-IE** | 685.67 | 1539.31 | 995.41 | 222.60 |
| **RA-CE** | 3313.23 | 5282.24 | 4453.31 | 369.76 |
| **RA-EA-IE** | **259.84** | **1501.51** | **509.90** | **272.46** |

Table 6 shows the distribution of the total working time for each worker for the best simulation (simulation which obtained the minimum objective function value), the worst simulation (simulation which got the maximum objective function value), and an average simulation for the particular case of the RA-EA-IE algorithm, for the Experiment 1 on a day with a low amount of deliveries. Note that we have included a simulation to illustrate the average case as close as possible to the actual average of 509.9s.

Figure 19 graphically illustrates the best result obtained for each algorithm, i.e., the distribution of the total working time for each worker in Experiment 1 for a day with a low number of packets to be delivered (see also Table 6). In this figure, it can be seen that both EA-IE and RA-EA-IE get a similar workload for all workers, slightly better for the RA-EA-IE, who work approximately the same amount of time. Also, the RA-IE algorithm obtains good performance. To get the workload balanced, algorithms RA-EA-IE, EA-IE, and RA-IE are capable for increase or decrease each type of time, i.e., $t_{Int}$, $t_{OW}$, $t_{Tra}$, $t_{Ext}$, and $t_{Ret}$, for each worker individually. Figures 19b, 19c and 19e show that there are di"erent worker types by their total working time distribution, i.e., there are workers with high $t_{Int}$ and $t_{Ext}$, that means they are workers with lots of packages to be delivered. Examples of this type of workers are Worker 7, Worker 8, and Worker 9 in Figure 19b, Worker 6 and Worker 10 in Figure 19c or Worker 8 and Worker 10 in Figure 19c. Other types of workers are those who have fewer delivery packages to be distributed, but those delivery points are located far away from each other, so the workers have higher $t_{Tra}$, or the delivery zone is far from the delivery center, so the workers have higher $t_{OW}$ and $t_{Ret}$, e.g., Worker 3 in Figure 19a, Worker 5 in Figure 19c or Worker 5 in Figure 19e.

Table 6: Example of the total working time for each worker and fitness function value of the algorithm RA-EA-IE in 3 cases: best, worst, and average.

|  | Best sim.(s) | Worst sim.(s) | Average sim.(s) |
|---|---|---|---|
| **Worker 1** | 8400,77 | 8641,69 | 8607,21 |
| **Worker 2** | 8574,54 | 8509,79 | 8630,26 |
| **Worker 3** | 8495,65 | 8586,95 | 8505,18 |
| **Worker 4** | 8652,36 | 8606,54 | 8636,85 |
| **Worker 5** | 8561,57 | 8518,37 | 8740,22 |
| **Worker 6** | 8439,33 | 8534,73 | 8695,30 |
| **Worker 7** | 8447,97 | 8783,66 | 8931,06 |
| **Worker 8** | 8651,17 | 8628,22 | 8739,29 |
| **Worker 9** | 8392,53 | 9404,14 | 9017,03 |
| **Worker 10** | 8515,79 | 8649,22 | 8710,57 |
| **Worker 11** | 8526,45 | 8708,59 | 8797,05 |
| **Worker 12** | 8447,71 | 7902,63 | 8764,94 |
| **Fitness F. value** | **259.84** | **1501.51** | **511.85** |

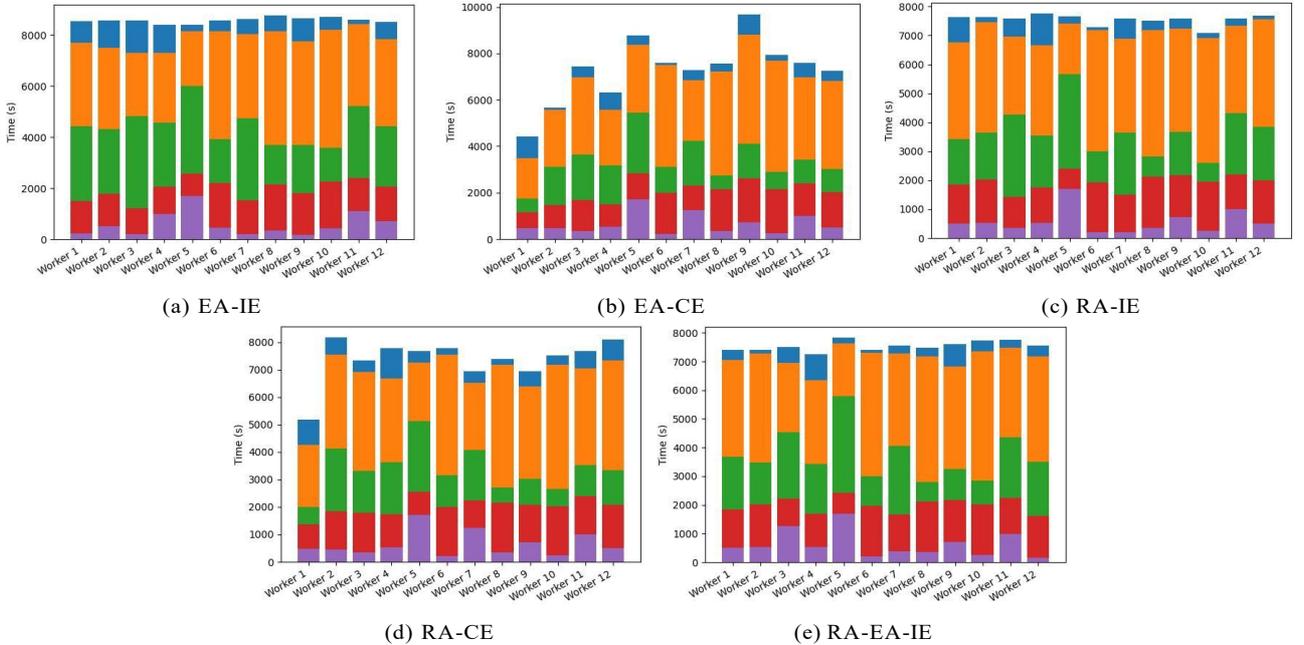

Figure 19: Distribution of the total working time for every worker for each algorithm on a type of day 1, the day with a low number of delivery packages.

Figure 20 shows the delivery assignment for each algorithm in Experiment 1, type of day 1. Each distribution point is colored by the color of the worker in charge of distributing the package, and the delivery center is represented by a red point (bigger than the other points). It is possible to see that the assignment of each delivery point to its corresponding worker is very similar between the algorithms that use Integer Encoding, i.e., the EA-IE algorithm, the RA-IE algorithm, and the RA-EA-IE algorithm so that the equality of performance discussed above can be appreciated. It is also possible to observe the di"erent types of workers mentioned above, e.g., the worker with many delivery addresses but close to each other (orange color in Figure 20a, in Figure 20c, and Figure 20e), or the worker with few but distant delivery points, also far from the delivery center (red color in Figure 20a, in Figure 20c and Figure 20e). Instead of that, the algorithms with circle encoding, i.e., EA-CE (see Figure 20b) and RA-CE (see Figure 20d), have low performance because they are assigning a similar number of delivery points without taking into account the point-to-point movement.

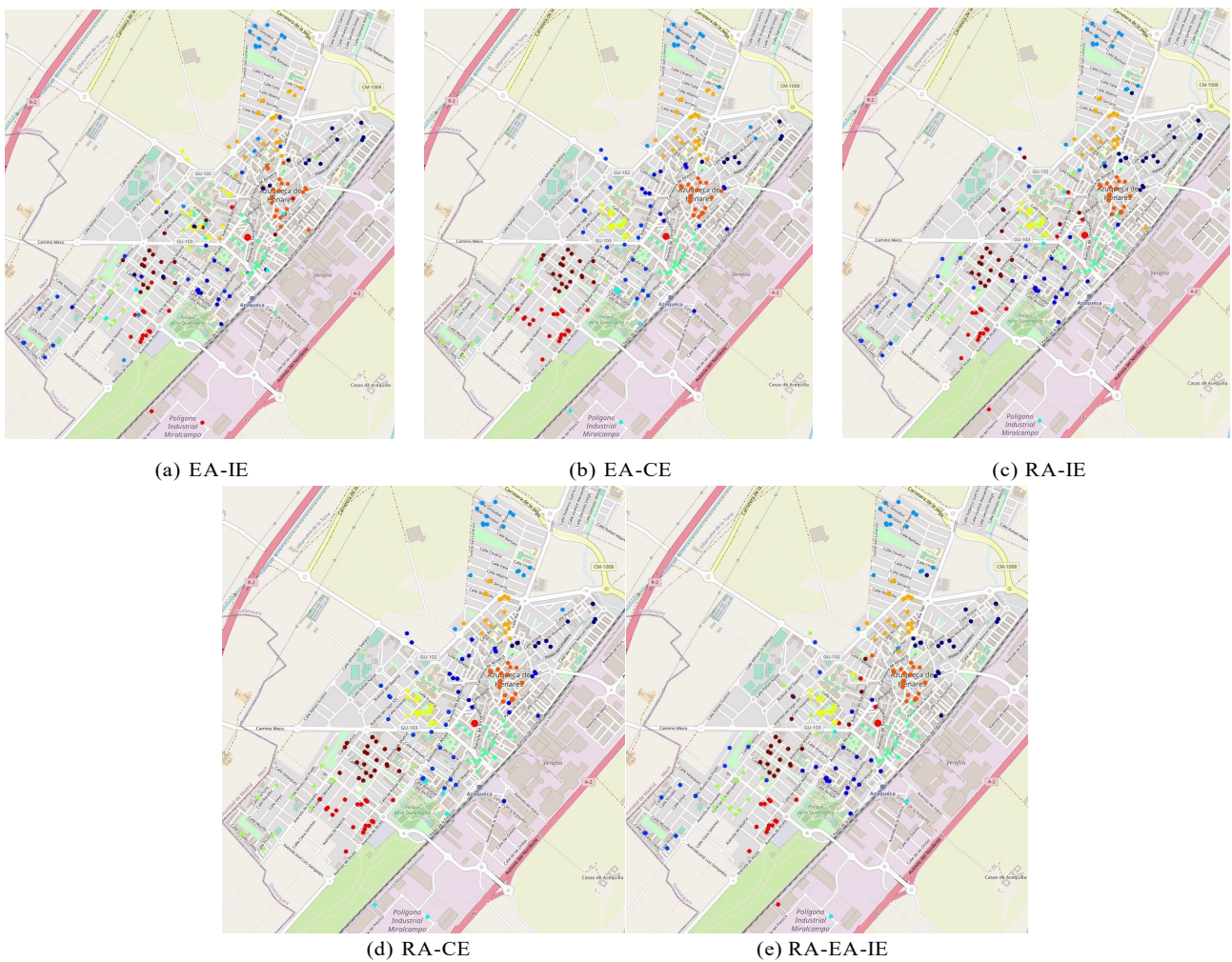

(a) EA-IE  (b) EA-CE  (c) RA-IE

(d) RA-CE  (e) RA-EA-IE

Figure 20: Packages delivery assignment solutions for the first type of day (low amount of packets to be delivery).

Table 7 shows the results obtained for Experiment 1, type of day 2 (a day with an average number of packages to be delivered). It can be seen that the best algorithm is again the RA-EA-IE, as in the case of a low number of packages to be delivered. On the simulation of this second day, the second-best algorithm in terms of performance of the objective function value is the RA-IE algorithm, ahead of the EA-IE approach. The reason for this result is the scenario's complexity, as the solution encoding vector is larger on a type 2 day than on a type 1 day, so the solution space exploration is more di!cult for an Evolutionary Algorithm. This problem does not happen in the RA-EA-IE algorithm because a part of the initial population is composed of solutions of the RA-IE algorithm, which enables the algorithm to start from a good solution. Finally, as in Day 1, the worst results are carried out by the algorithms that use circle encoding.

Table 7: Experiment 1 fitness function statistical values for each algorithm considered for a day with an average number of packets to be distributed (type of day 2).

|          | Min.(s)  | Max.(s)  | Mean(s)  | Std.(s) |
|----------|----------|----------|----------|---------|
| **EA-IE**    | 2390.36  | 5181.04  | 3669.32  | 794.96  |
| **EA-CE**    | 5014.98  | 10578.3  | 7481.98  | 2578.50 |
| **RA-IE**    | 1316.90  | 2987.76  | 2146.59  | 414.84  |
| **RA-CE**    | 3025.40  | 9484.76  | 5458.63  | 1971.60 |
| **RA-EA-IE** | **338.65** | **1868.35** | **696.26** | **368.86** |

Figure 21 graphically shows the best solutions obtained by the di"erent algorithms in this problem. It is possible to see that the RA-EA-IE algorithm gets a good workload balance, and the workers work the same amount of time, approximately. Again, to get the workload balanced, the algorithm RA-EA-IE is capable of increasing or decreasing each type of time. In Figure 21e, it can also be seen that there are di"erent types of workers depending on their total time distribution, e.g., workers with a high number of deliveries in small areas, workers with a low amount of deliveries on large areas, among others. The RA-IE algorithm and the EA-IE algorithm achieve good performance by doing the same workload balancing as the RA-EA-IE algorithm, but it can be seen that the performance is worse than that of the RA-EA-IE algorithm. The EA-CE and the RA-CE algorithms do not achieve acceptable and balanced solutions.

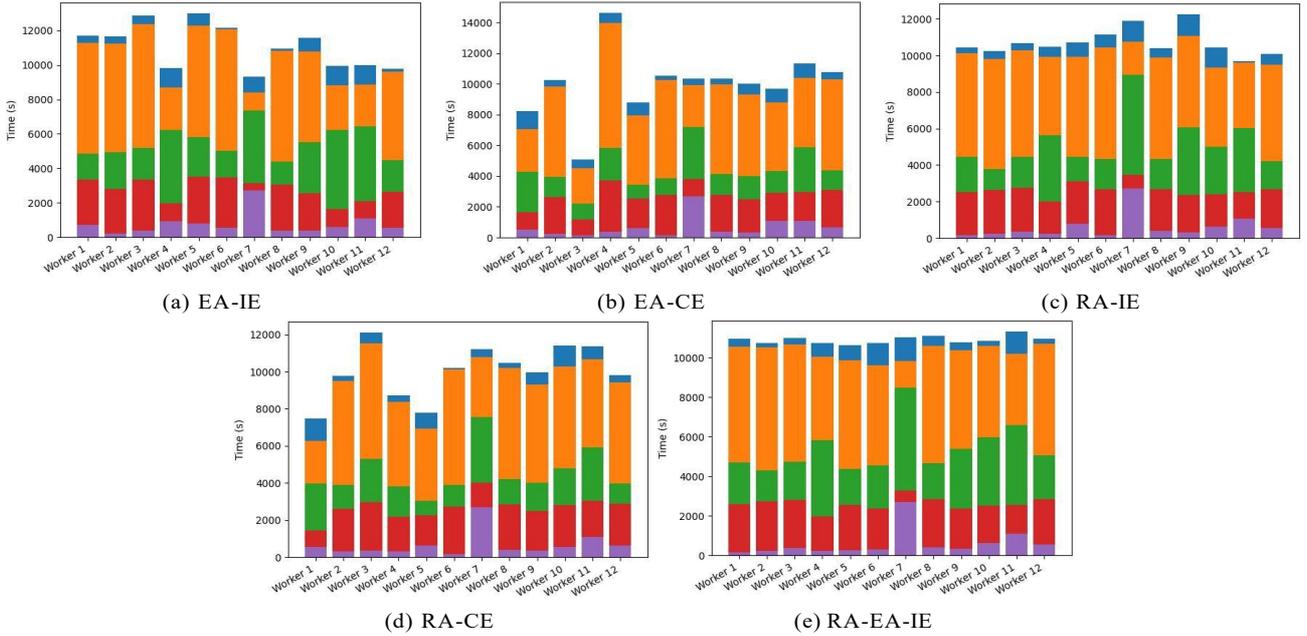

Figure 21: Distribution of the total working time for every worker for each algorithm on a type of day 2, the day with an average number of delivery packages.

Figure 22 shows the second type of day delivery package geographical distribution in Azuqueca de Henares for each algorithm. Each distribution point is colored by the color of the worker in charge of distributing the package, and the delivery center is again represented by a red point bigger than the other points. It can be seen in the RA-EA-IE algorithm (see Figure 22e) that there are workers with close delivery points who are assigned a large number of points because their work distribution time is mainly based on delivery (e.g., yellow or orange workers), or there are workers with distant delivery points who are assigned a reduced number of points because their work distribution time is mainly based on point-to-point movement (e.g., red or light green workers). In algorithms with integer encoding, the distribution is similar among them and the changes between them are minimal. As seen on type of day 1, the algorithms with circle encoding, i.e., EA-CE and RA-CE, have low performance because they assign a similar number of delivery points without taking into account the point-to-point movement.

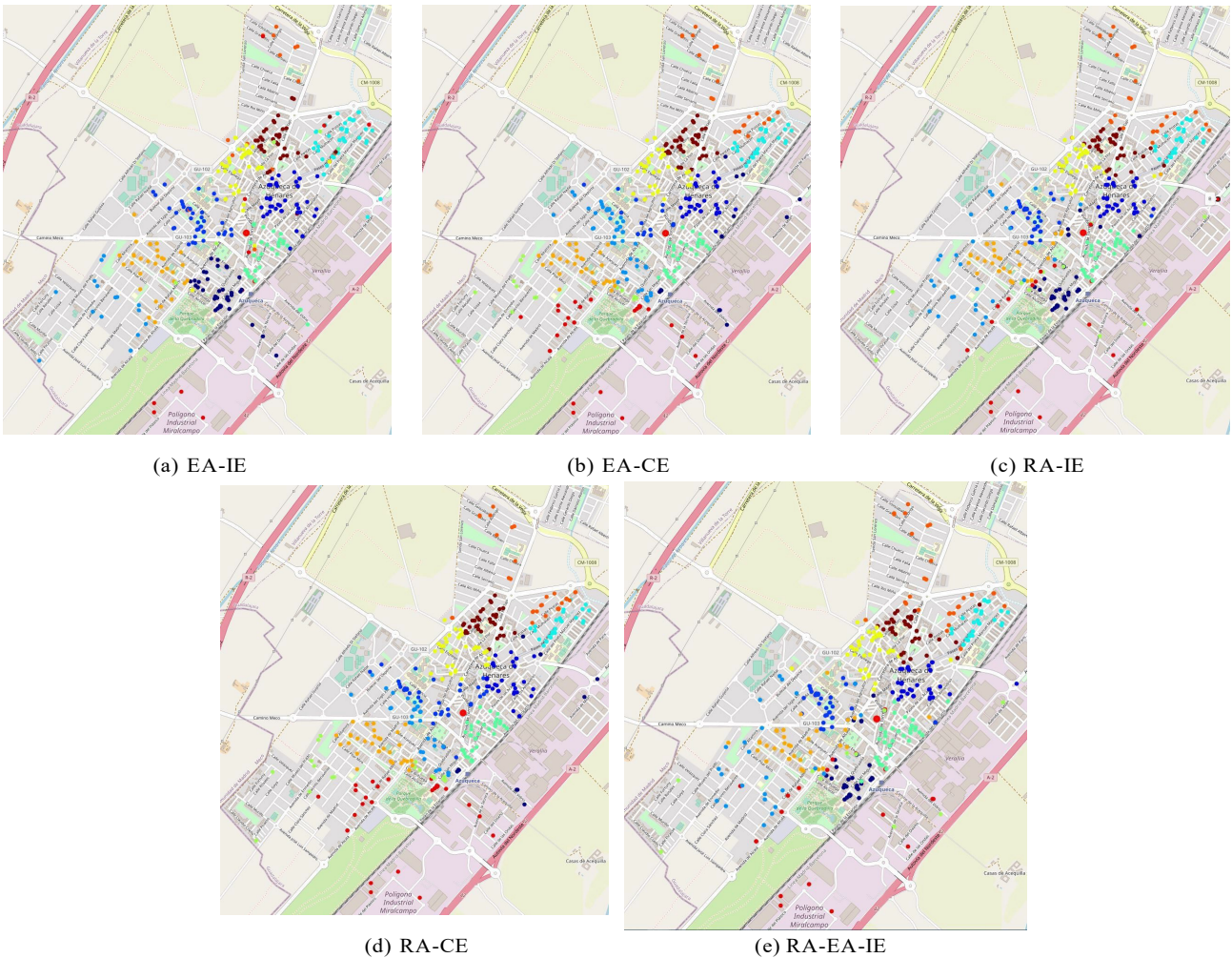

(a) EA-IE    (b) EA-CE    (c) RA-IE

(d) RA-CE    (e) RA-EA-IE

Figure 22: Packages delivery assignment solutions for the second type of day (average delivery).

Finally, Table 8 shows the results obtained for the third type of day, a day with a high load of package deliveries, for each algorithm presented. It can be seen that the results are worse than in the first and second cases as the complexity of the scenario has increased. The scenario changes from having 240 packages and 12 workers on day type 1 and 392 delivery packages and 12 workers on day type 2, to having 628 delivery packages and 13 workers, so the solution exploration space becomes much larger, as shown in Table 1. This situation is reflected in the performance of all the algorithms. Although the results are worse than for the other scenarios, it can be seen that the performance order for the algorithms is the same. The best performance is achieved by the RA-EA-IE algorithm in all the statistical objective function values (i.e., the best simulation value, the worst simulation value, the mean of all simulation values, and the standard deviation). As happened on day type 2, the second algorithm with better performance is the RA-IE algorithm, followed by the EA-IE algorithm. The EA-IE algorithm obtains similar results as the RA-CE algorithm in the fitness function mean value for all the simulations and is worse in the minimum value of the fitness function. So EA-IE is a more stable algorithm in this third type of day of simulation, but there are some simulations in which the RA-CE algorithm performs better. Note that, as the scenario becomes complex, circle encoding algorithms also degrade in performance.

Figure 23 graphically shows the time distribution for each worker for each algorithm for the third type of day package distribution in Azuqueca de Henares. It can be seen graphically that the best algorithm in terms of workload balance between workers is carried out by the RA-EA-IE algorithm (see Figure 23e), which obtains the same amount of work time for each worker, approximately, as seen in Table 8. The RA-IE algorithm gets a good workload balance for almost all workers, but there is one worker whose workload is high, making the fitness function worse (see worker 2 in Figure 23c). As shown in Table 8, the EA-IE algorithm is the third algorithm in terms of performance, and it can be seen in Figure 23a that there are two workers with low workloads (Worker 10 and Worker 12) and there are two workers with high workloads (Worker 3 and Worker 13). Again, the RA-CE and EA-CE algorithms are the worst algorithms due to the type of encoding.

Table 8: Experiment 1 fitness function statistical values for each algorithm considered for a day with a high number of packets to be distributed (type of day 3).

|          | Min.(s)  | Max.(s)   | Mean(s)   | Std.(s)  |
|----------|----------|-----------|-----------|----------|
| **EA-IE**    | 6393.75  | 8580.47   | 7490.42   | 616.68   |
| **EA-CE**    | 7489.54  | 15142.50  | 12746.52  | 1984.34  |
| **RA-IE**    | 2557.64  | 4968.63   | 3581.23   | 642.65   |
| **RA-CE**    | 3678.56  | 11829.06  | 8389.22   | 1620.71  |
| **RA-EA-IE** | 1415.78  | 3430.77   | 2035.21   | 436.98   |

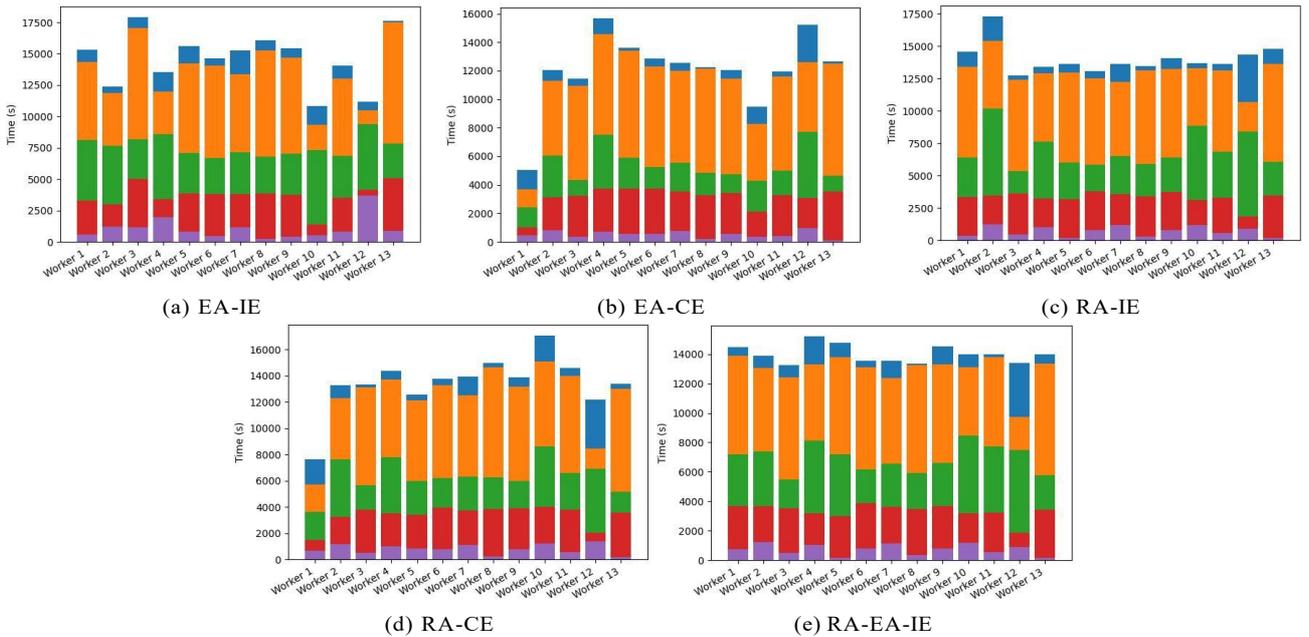

(a) EA-IE    (b) EA-CE    (c) RA-IE

(d) RA-CE    (e) RA-EA-IE

Figure 23: Distribution of the total working time for every worker for each algorithm on a type of day 3, the day with a high number of packages to be delivered.

Finally, Figure 24 shows the third type of day package delivery assignment in Azuqueca de Henares for each algorithm. Each worker's assigned delivery is defined by a di"erent color. The distribution of the asignation follows a similar pattern as in the first and second types of days, with the RA-EA-IE obtaining the most balanced assignment.

*4.3. Results for Experiment 2*

We now describe the second carried out experiment to illustrate the performance of the best proposed approach in Experiment 1 (RA-EA-IE) for this problem of operational human resources workload balancing in a last-mile urban delivery system. Thus, Experiment 2 consists of the RA-EA-IE algorithm performance analysis for di"erent days in a row. This experiment is carried out to demonstrate its good performance in diverse package delivery situations and with distinct numbers of workers, as shown in Table 2. Simulation parameters and hyperparameters are also the same as in Experiment 1 (see Section 4.2) and can be consulted in Table 3 and Table 4 respectively.

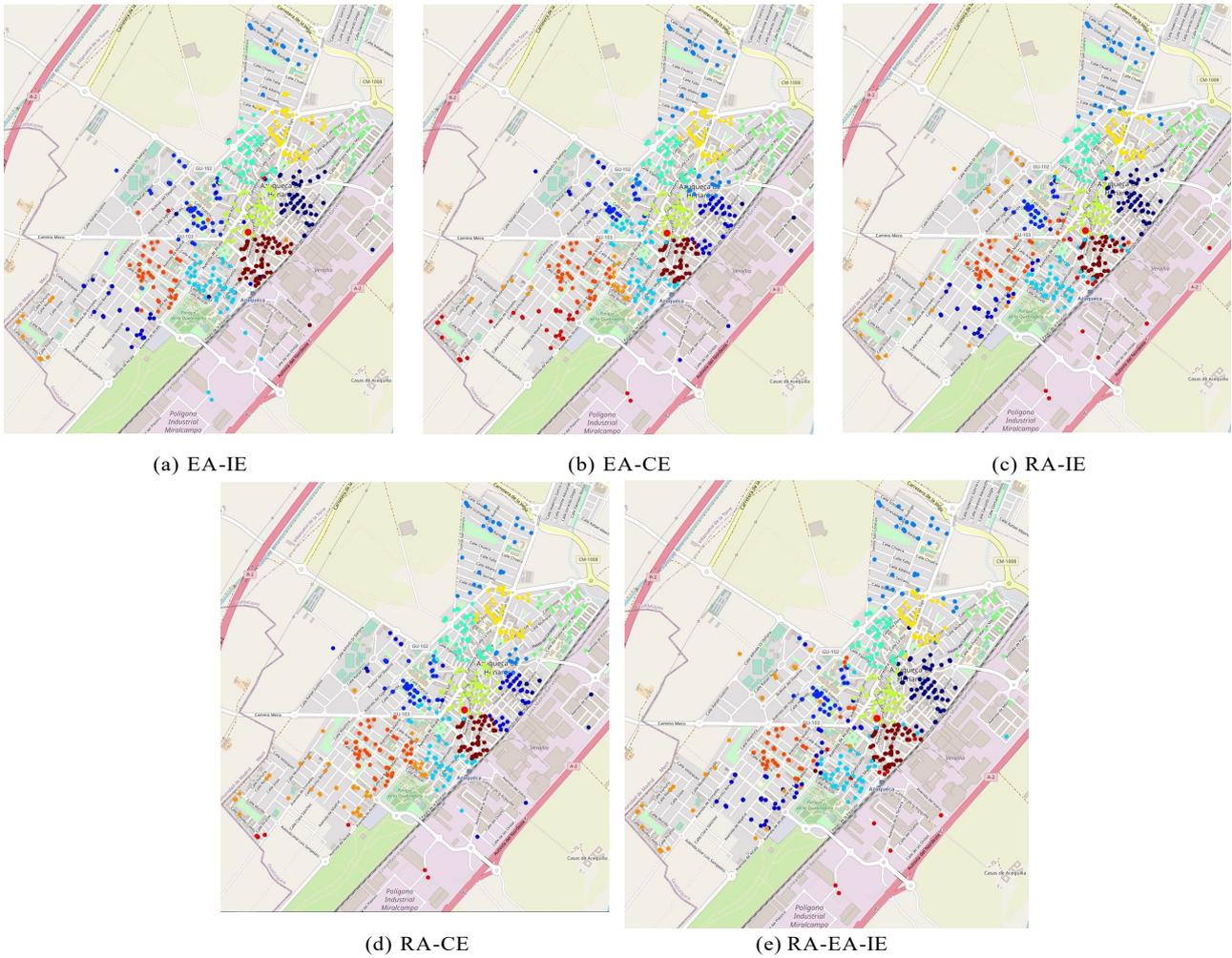

Figure 24: Packages delivery assignment solutions for the third type of day (high load of deliveries).

Table 9 shows the objective function statistical values using the RA-EA-IE algorithm for all days considered in all the simulations. It is possible to see that there is a correlation between the results obtained and the total number of packages to be delivered (see Table 2) for each day. The greater the number of delivery packages to be delivered (i.e., the more complicated the scenario), the worse the results reached by the RA-EA-IE algorithm, both in terms of the mean value and the standard deviation of the simulations.

We can also compare days with a similar number of packages to be delivered but di"erent numbers of workers in the system to extract further conclusions about the RA-EA-IE algorithm performance.

The first comparison is between day 1 (with 715 delivery packages and 13 workers) and day 9 (with 676 delivery packages and 11 workers). It can be observed that the number of delivery packages is similar, but the number of workers is di"erent, making day 1 a more complicated scenario. The results show that day 9 is performing better than day 1 with respect to the mean value and standard deviation. Another case of comparison is between days 2 and 3 and days 10 and 11, where the number of delivery packages on these days is similar, but days 2 and 3 form a less complex scenario than days 10 and 11 because of the number of workers. It is possible to see that the performance comparison is similar to the first case, where the algorithm performs better in less complex scenarios than in complex scenarios in terms of mean value and standard deviation. The final comparison is between day 5 and day 6, where day 5 is a less complex scenario than day 6 but only gets better results in the minimum and the mean value of the fitness function for all the simulations. This result may be due to the homes and stores geographical distribution, which makes day 5 a more complicated scenario. Figure 25 shows the obtained objective function values for each day for all the simulations, as in Table 9. It is possible to see the correlation between Figure 18 and Figure 25, i.e., in general, if the scenario is more complex, the results are worse in mean and standard deviation. From the above comparisons and looking at Figure 25, day 9 has a better performance than day 1 in terms of mean, median, and standard deviation, as same as days 2 and 3 against days 10 and 11. Between days 5 and 6, it is possible to observe that day 5 has a better performance on the median value of the fitness function (yellow line), but in the rest of the statistical data, it is worse than day 6.

Figure 26 shows the total working distribution time for each worker in the best simulation for each day in Experiment 2 using the RA-EA-IE algorithm. It is possible to see that the algorithm always obtains a good performance of workload balancing among all the workers for each day, even with di"erent numbers of workers (some days have 11 workers, some days have 12 workers, and some days have 13 workers to cover the same geographical area). To obtain this overall workload balance, the algorithm balances the di"erent time contributions, i.e., $t_{Int}$, $t_{OW}$, $t_{Tra}$, $t_{Ext}$, and $t_{Ret}$. It means that there are workers whose total working time is defined by a high number of packages to be delivered, but the geographical distribution of the delivery packages is close among them, e.g., worker 4 on day 5 (see Figure 26e). Other workers have a low number of delivery packages to be distributed, but the geographical distribution of the delivery packages is far away among them, so they have a high $t_{Tra}$, e.g., worker 5 on day 1 (see Figure 26a). Other workers have their initial work zone far away from the delivery center, so they have high $t_{OW}$ and $t_{Ret}$ instead of others, e.g., worker 3 on day 11 (see Figure 26k). In general, the algorithm can manage these cases to obtain the best solution for the human resources workload balancing in a last-mile urban delivery system problem.

Table 9: Experiment 2 fitness function statistical values using RA-EA-IE algorithm for all days considered.

|  | Min.(s) | Max.(s) | Mean(s) | Std.(s) |
|---|---|---|---|---|
| **Day 1** | 845.88 | 2442.39 | 1529.57 | 437.53 |
| **Day 2** | 308.49 | 1713.30 | 691.58 | 297.59 |
| **Day 3** | 233.75 | 1397.86 | 573.28 | 240.44 |
| **Day 4** | 546.24 | 2385.91 | 1351.84 | 579.62 |
| **Day 5** | 475.37 | 2585.49 | 1309.30 | 570.79 |
| **Day 6** | 745.61 | 2315.44 | 1269.86 | 391.20 |
| **Day 7** | 399.51 | 2023.07 | 1030.27 | 523.69s |
| **Day 8** | 385.00 | 1733.26 | 921.55 | 395.19 |
| **Day 9** | 556.43 | 1643.64 | 1046.10 | 289.54 |
| **Day 10** | 404.82 | 1650.82 | 928.51 | 350.20 |
| **Day 11** | 411.97 | 2014.93 | 866.01 | 484.12 |
| **Day 12** | 928.43 | 3481.10 | 1545.01 | 558.37 |

*4.4. Further results, analysis and discussion*

In this section, we discuss the experiments carried out and the results obtained involving the Classical Clustering Algorithms (Section 3.1), i.e., k-means clustering algorithm, Gaussian Mixture Models, Balance Iterative Reducing and Clustering using Hierarchies (BIRCH), Agglomerative Hierarchical Clustering and Spectral Clustering. Without loss of generality, these algorithms have been tested in the second type of day (day 2 of Experiment 1), with an average number of packages and workers (see Table 1). The results obtained have been compared with the algorithms and methods described in Section 3, i.e., EA-IE, EA-CE, RA-IE, RA-CE, and RA-EA-IE. As in other experiments, we run 30 simulations of each algorithm to obtain statistically significant results, and we have calculated several statistical parameters, e.g., minimum value, maximum value, mean value, and standard deviation of the fitness function cost value for all the algorithms. The general parameters for each algorithm are the same as in the previous experiments (see Table 3).

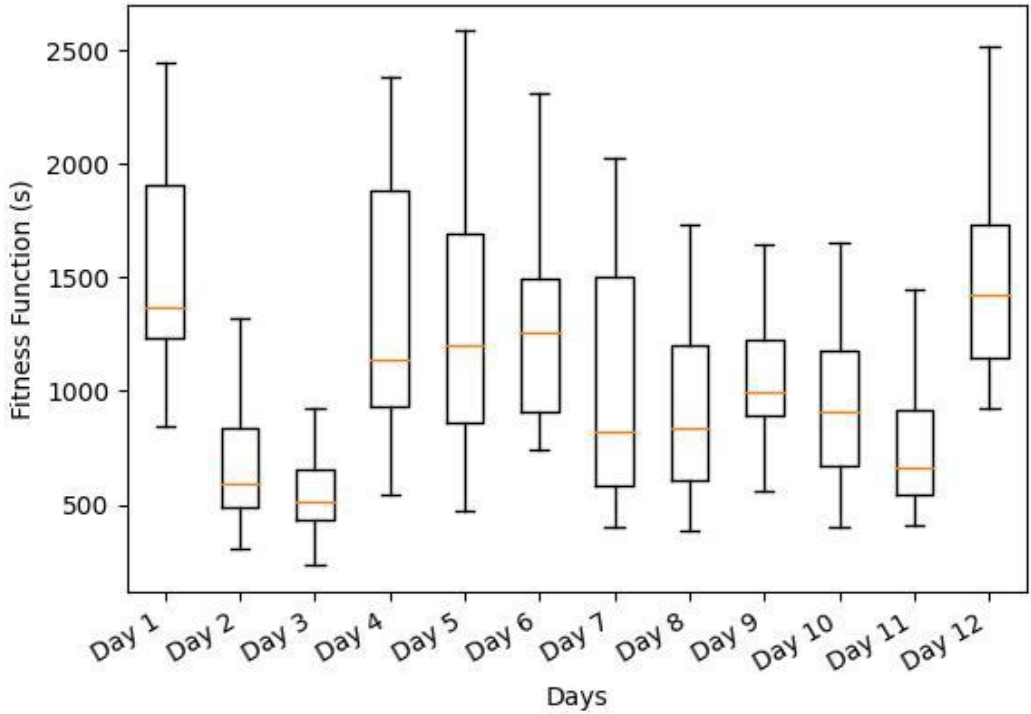

Figure 25: Experiment 2 objective function statistical values for every day considered.

The results obtained by the Classical Clustering Algorithms are shown in Table 10. It is possible to see that the results of Classical Clustering Algorithms on their own are, in general, poor, indicating the need to develop new algorithms and methods. It also can be seen that Spectral Clustering sometimes provides acceptable solutions, but in other cases, the solutions obtained are poor, so we can say that the algorithm's convergence is not very good. The k-means algorithm is the more stable among Classical Clustering Algorithms, but the results are poor compared to the algorithms developed in this work, as expected. Agglomerative Clustering and BIRCH provide sturdy solutions, but the cost value of the fitness function is worse than that of other algorithms.

Table 10: Clustering Algorithms fitness function statistical values for each algorithm considered for a day with an average number of packets to be distributed (type of day 2).

|  | **Min.**(s) | **Max.**(s) | **Mean**(s) | **Std.**(s) |
| --- | --- | --- | --- | --- |
| **k-Means** | 7818.99 | 8310.79s | 8105.01 | 128.55 |
| **Gaussian Mixture** | 7705.67 | 13962.58 | 10784.27 | 1602.77 |
| **Agglomerative Clustering** | 10897.76 | 11506.93 | 11237.78 | 179.85 |
| **BIRCH** | 10517.42s | 11128.14 | 10930.49 | 180.39 |
| **Spectral Clustering** | 2878.80 | 11853.53 | 6195.46 | 1747.56 |

Figure 27 shows the obtained objective function values for each Clustering Algorithm, i.e.

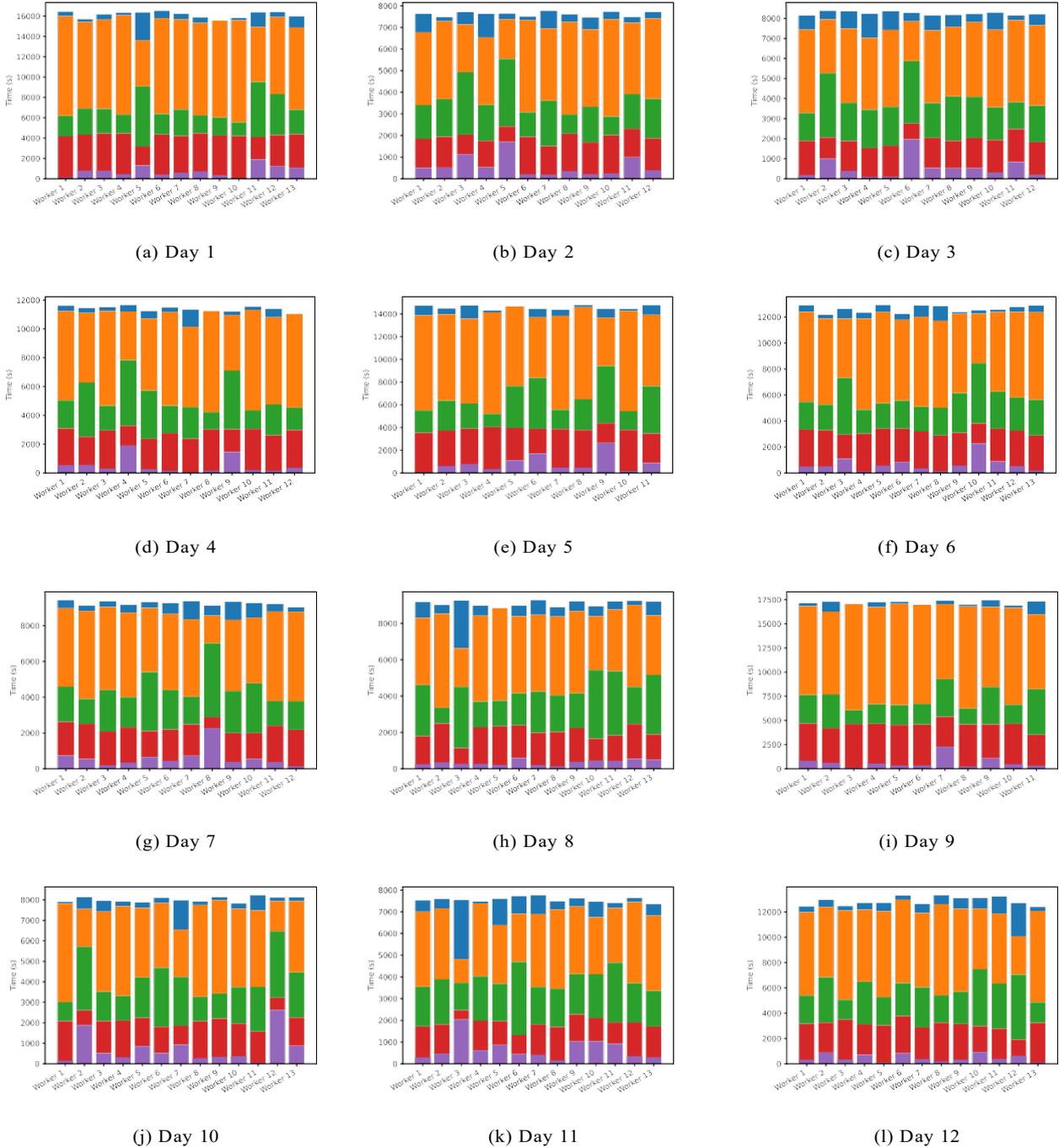

Figure 26: Experiment 2 worker distribution time for each day's best simulation.

k-means (KM), Gaussian Mixture (GM), Agglomerative Clustering (AC), Balance Iterative Reducing and Clustering using Hierarchies (BIRCH) and Spectral Clustering (SC), as in Table 10. It is also easy to see that in some simulations, Spectral Clustering obtained quite good performance on its own, though the variance of the solutions obtained is high, indicating that the algorithm performance in some other simulations is not so good. These results justify the need to develop an alternative approach, such as the proposed Recursive Algorithm with Spectral Clustering initialization and integer encoding, against the Recursive Algorithm with k-means initialization (see Section 3.4).

Figure 28 shows the distribution of the total working time for each Classical Clustering algorithm for each worker in the best solution when the number of delivery packages to be distributed is an average value. It is possible to see that Spectral Clustering is the only algorithm capable of getting a good workload balance between all workers, and the performance of the other algorithms is poor. The clusters formed by the algorithms with poor performance, i.e., k-means, Gaussian Mixture Models, Agglomerative Clustering, and BIRCH, are built using a distance metric, so the cluster's shape is relatively close to a circle, as can be seen in Figure 29. As we demonstrated in Experiment 1 (see Section 4.2), the algorithms that use circle encoding get a poor performance due to the low flexibility of this kind of encoding, and the shape of the clusters formed is quite similar to the shape of the clusters formed by the Classical Clustering algorithms, except Spectral Clustering. That is the reason for the bad results with these clustering algorithms. The good results of the Spectral Clustering algorithm are due to the flexibility of the shape of the clusters formed.

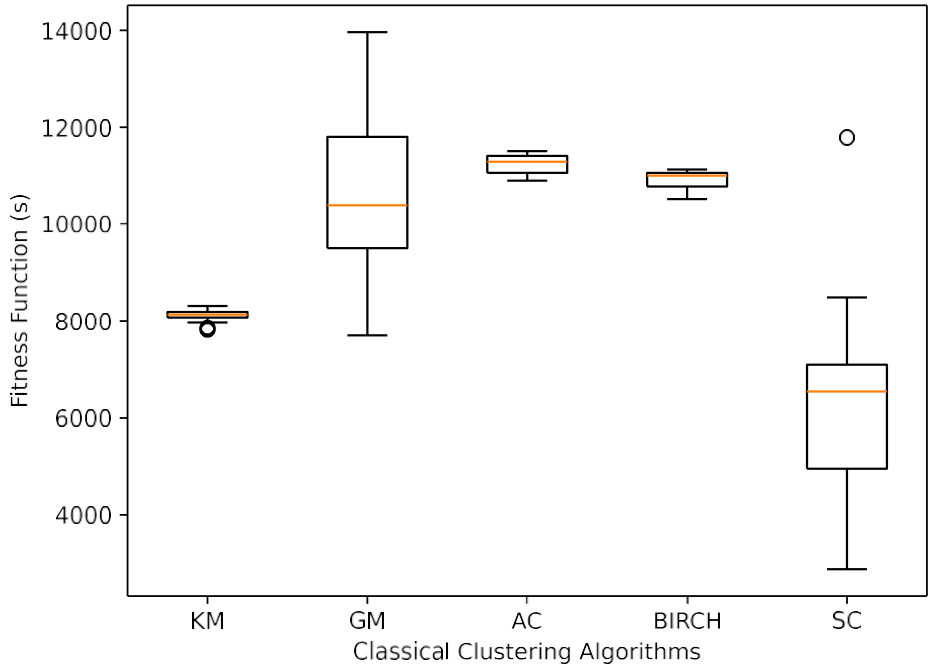

Figure 27: Classical Clustering Algorithms objective function statistical values for a day with an average number of packets to be distributed (type of Day 2).

Furthermore, these results of the Classical Clustering approaches suggest the feasibility of using the SC approach instead of k-means in the proposed method. To evaluate this possibility, we have tested the proposed algorithm with the best performance in the problem at hand (the RA-EA-IE, see Section 3.6), but substituting the k-means algorithm with the SC clustering. Table 11 shows the results obtained for the three days considered in Experiment 1 (Section 4.2). As can be seen, the results obtained by the RA-EA-IE algorithm based on SC clustering are, in general, worse than those by the proposed RA-EA-IE with k-means: in the type of day 1 and type of day 2 cases (low and average number of packets to be delivered), the RA-EA-IE with k-means obtains better results than the SC in all the metrics considered, obtaining always better solutions and with less variability in the algorithm's convergence. In the type of day 3 (high number of delivery packets), the RA-EA-IE with SC can obtain the best solution overall, with an excellent result of 637 seconds in the objective values, versus a best result of 1415 seconds obtained by the RA-EA-IE with k-means. The rest of the metrics indicate that the algorithm convergence using the k-means is much more stable, and the solutions obtained show much less variability than those using the SC.

Figure 30 visually shows these results, with green boxes belonging to the RA-EA-IE with k-means and the blue ones to the RA-EA-IE with SC. It is straightforward that, in general, the performance of the algorithm with the k-means is better, but the SC can obtain punctual solutions of high quality, at least in the day with the highest number of deliveries programmed. The reason for this seems evident by looking at Figure 29 clusters: While k-means (and also other clustering approaches such as GM, AC, and BIRCH) provide well-defined, compact, and stable clusters, the clusters obtained by the SC approach are not compact at all. It seems that this behavior can lead to more diverse centroids (initial points for the RA-EA-IE), so the variability in the solutions obtained by the ensemble algorithm is high. Sometimes, this variability in the RA-EA-IE initialization may lead to finding acceptable solutions to some particular problems.

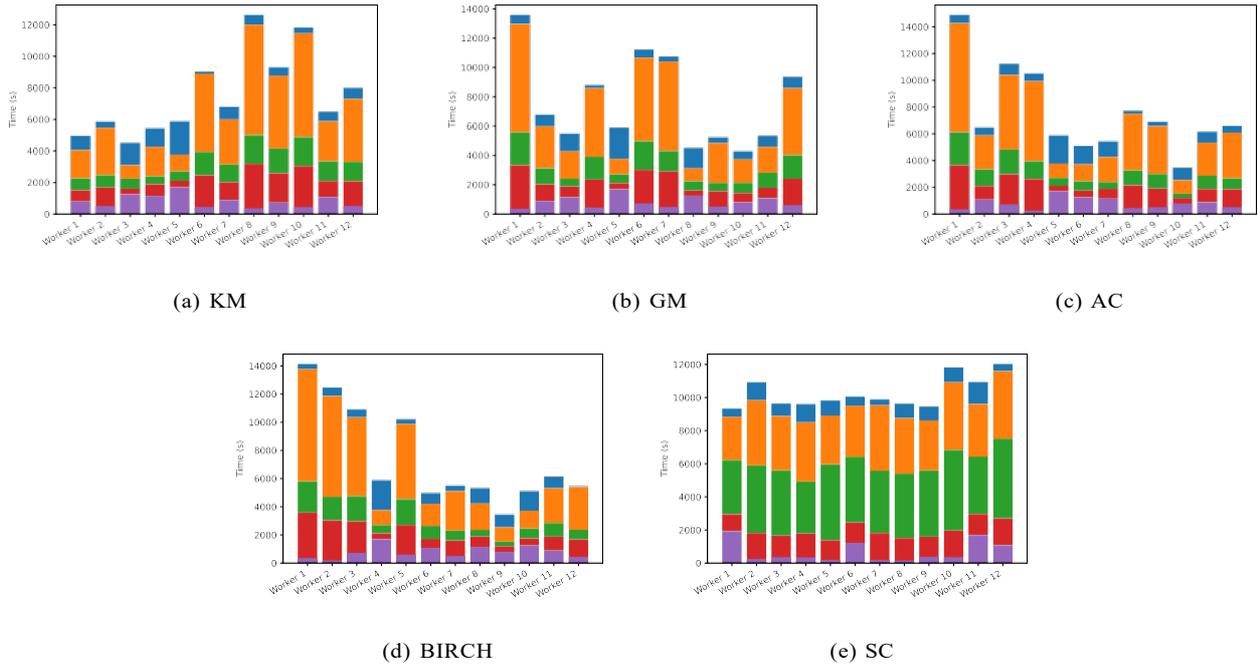

Figure 28: Distribution of the total working-time for each Classical Clustering algorithm in the best solution on for day 2 (day with an average number of packages to be delivered).

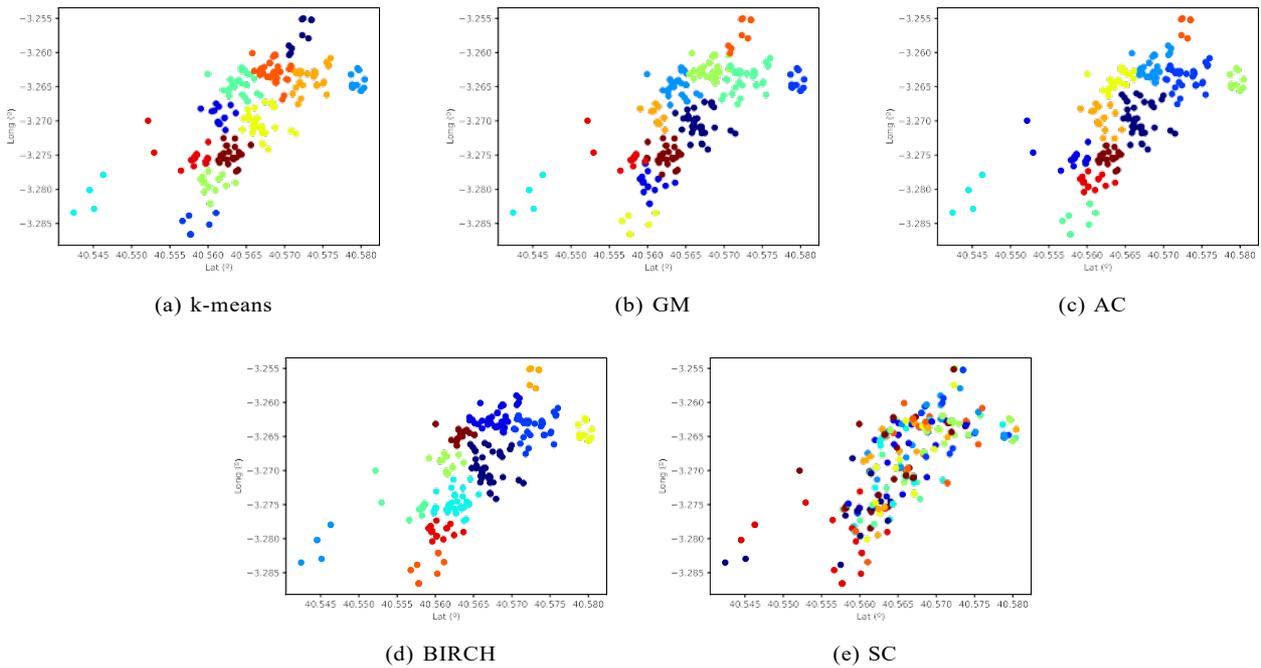

Figure 29: Distribution of the designated clusters for each Classical Clustering algorithm in the best solution on a type of day 2 (average number of packages to be delivered).

Table 11: RA-EA-IE fitness function statistical values substituting the k-means algorithm by the Spectral Clustering (SC) algorithm.

|  |  | Min. (s) | Max. (s) | Mean(s) | Std.(s) |
|---|---|---|---|---|---|
| Day 1 | RA-EA-IE (SC) | 380.67 | 2081.51 | 969.94 | 486.23 |
|  | RA-EA-IE (k-means) | 259.84 | 1501.51 | 509.90 | 272.46 |
| Day 2 | RA-EA-IE (SC) | 405.59 | 1972.73 | 1121.05 | 409.42 |
|  | RA-EA-IE (k-means) | 338.65 | 1868.35 | 696.26 | 368.89 |
| Day 3 | RA-EA-IE (SC) | 637.14 | 4026.22 | 2294.52 | 835.69 |
|  | RA-EA-IE (k-means) | 1415.78 | 3430.77 | 2035.21 | 436.98 |

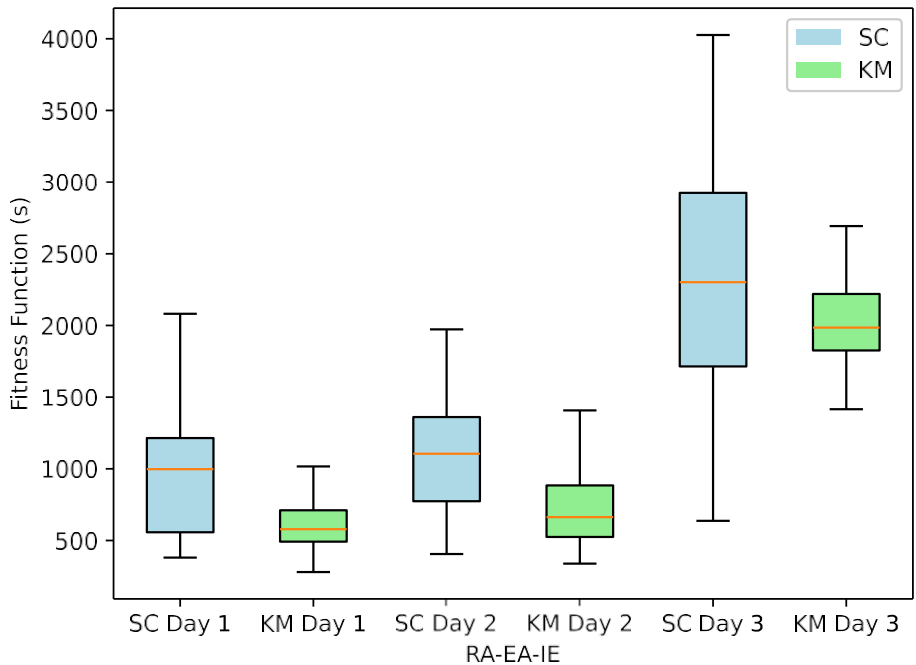

Figure 30: Comparison of objective function statistical values between k-means and SC initialization in the RA-EA-IE algorithm.

We can further discuss the specific performance of the SC in the RA-EA-IE algorithm by depicting the best result obtained in terms of delivery times and comparing it with the best solution found with the RA-EA-IE with k-means for this problem. Thus, Figure 31 shows the comparison in the distribution of working time for each worker between the RA-EA-IE algorithm using k-means or SC as initialization algorithm. As mentioned, these figures correspond to the best simulation for each algorithm on day 3 (the day with a high number of packages to be delivered). It is possible to see that the SC algorithm can obtain a better performance in terms of the objective function cost value (see Equation 7). Specifically, due to the objective function considered, it means that the di"erence between the worker with the highest working time and the worker with the lowest working time is better in the RA-EA-IE with the SC algorithm than in the RA-EA-IE with k-means. A closer analysis of the solutions obtained shows that the total working time for each worker is higher using the SC algorithm than with the k-means algorithm, so the improvement in balancing the working time is obtained at the cost of increasing the total delivery time. The explanation of these results can be better observed in the green color bar for each worker in Figure 31, which corresponds to the Travel Time ($t_{Tra}$). The clusters formed by the SC algorithm have more flexibility, whereas the clusters formed by the k-means algorithm have a shape more similar to a circle, reducing the flexibility of the solutions but getting better total working time for each worker, as can also be observed in Figure 32. These results suggest that the objective function should include a term for minimizing the total working time in addition to the minimization of the workload di"erences.

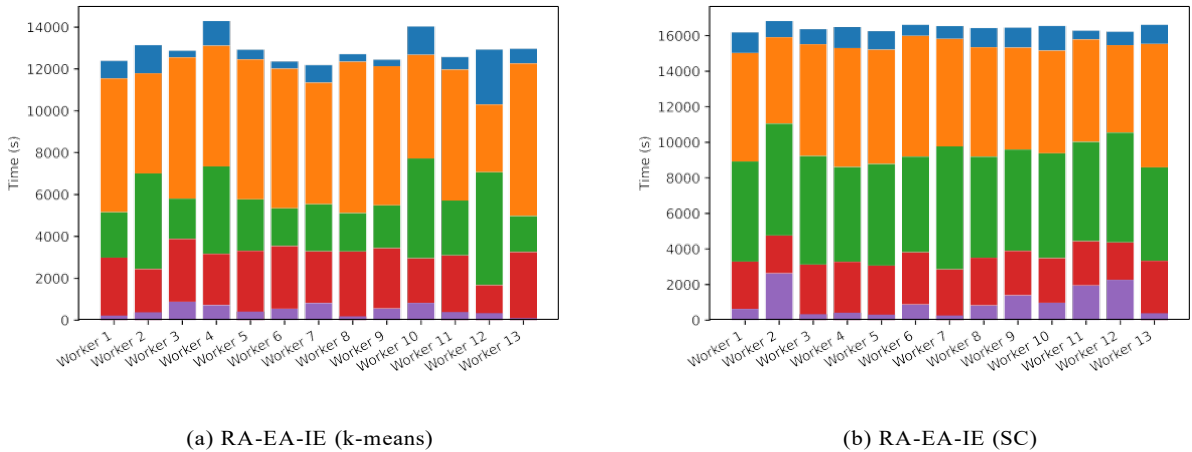

(a) RA-EA-IE (k-means)   (b) RA-EA-IE (SC)

Figure 31: Comparison in the distribution of the total working-time for RA-EA-IE algorithm with k-means and SC initialization in the best solution for day 3 (day with a high number of packages to be delivered).

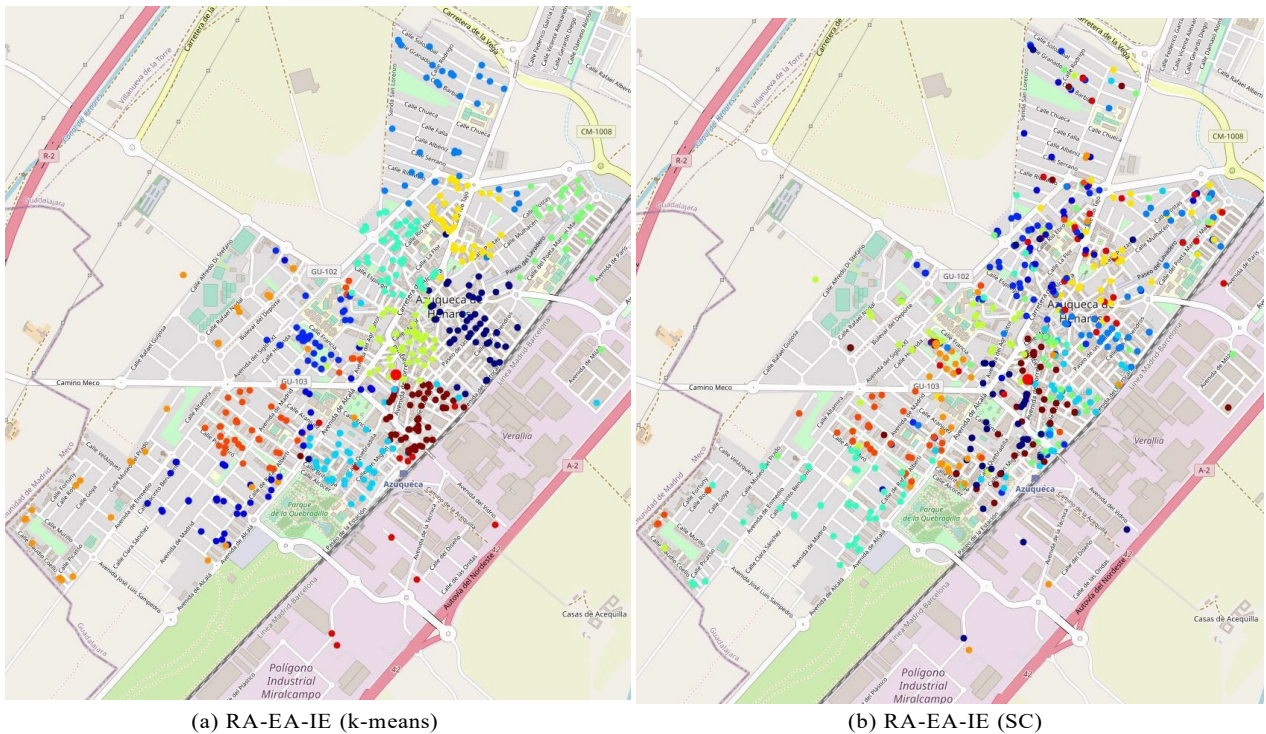

(a) RA-EA-IE (k-means)   (b) RA-EA-IE (SC)

Figure 32: Comparison in the packages delivery assignment solutions for RA-EA-IE algorithm with k-means and SC initialization in the best solution for day 3 (day with a high number of packages to be delivered).

## 5. Conclusions

In this paper, we have proposed a multi-algorithm approach for optimizing daily human resources workload balancing in package delivery systems. The problem tackled is defined with a set of delivery points and a predefined number of workers as input, and the expected output is the assignment of packages to workers in a way that each one works a similar time per day. Specifically, di"erent versions of Evolutionary Algorithms and recursive heuristics based on k-means initialization have been proposed. We have considered two di"erent types of encodings which a"ect the algorithms' structure and operators. The first encoding proposed is an integer encoding, where the solution consists of a vector with equal size to the number of delivery points in the city, and each element of that vector corresponds to the assigned worker to the delivery. The second encoding proposed is a circle encoding, where the solution consists of a vector representing a circular area in which every worker must deliver.

We have successfully illustrated the performance of the proposed approach in a real-world problem of human resource balancing in an urban last-mile package delivery workforce operating at Azuqueca de Henares, Guadalajara, Spain. The results obtained have shown that, in this particular problem, integer encoding gets better results than circle encoding due to the flexibility provided by integer encoding. We have also shown that the hybrid approach formed by an Evolutionary Algorithm with Recursive Assignment and Integer Encoding (RA-EA-IE) is the approach that obtained the best results overall algorithms tested in this paper. The results obtained by the RA-EA-IE showed an excellent balance in di"erent scenarios with low, average, and high working loads. The multi-algorithm approach presented here can be modified and extended to work in alternative logistics-related problems, such as assignment tasks [41, 42] or allocation problems [43, 44], among others.

Regarding the limitations of the proposed model, maybe the most important one is the fact that we do not work with exact delivery working times, but we model them using a Travelling Salesman Problem (TSP) that is an NP-hard problem itself. Thus, it is not even possible to obtain an exact solution for the TSP model considered, given the amount of data that we are working with. Therefore, the delivery working time is calculated using a heuristic algorithm (a Local Search Algorithm), so the total delivery working time used is an approximation solution for a model. Another limitation of this problem in general, and of the proposed methods and algorithms in particular, is the convergence of the algorithms when the size of the data scales. For example, in a considerable city, where the number of delivery packages and the number of workers is large, the computation time of the proposed approach may also be very high.

Finally, we comment on the future lines of research that this work may open: new hybrid algorithms implementation using alternative modern metaheuristic algorithms deserves special mention since it is probable that the improvement margin for this problem is still principal, especially for huge cities cases. Also, in close connection to the limitations outlined above, we would like to explore the integration of a better model for obtaining delivery working times (currently the TSP). Hybridizing the proposed approach with real routing applications is feasible, but it also has di"erent issues and other limitations. In this context, note that any improvement in the TSP solving (both in terms of accuracy or computation time) could help obtain better solutions to the daily human resources workload balancing problem tackled. Finally, a future line of research concerning the problem definition is the possibility of extending the worker typologies and including their specific delivery system (walking, car, motorcycle, among others) in the optimization. Thus, the idea would be to try to optimize the workload and the total working time of all the workers, with new constraints regarding the worker typology and with alternative problem restrictions. Another possible improvement is to redesign the objective function to take into account the total working time of the sta" in addition to the workload di"erences.

**Acknowledgements**

This work is supported by the Correos – UAH Innovation Research Chair. We would like to express our special gratitude to M. Climent, S. Villoria, E. Meruelo, and G. Gosálbez, from the Strategy Department, and P. Garc´ıa and J. Pozas from the Territory Management Department of Correos for supervising the work presented in this paper. The research has also been partially supported by the project PID2020-115454GB-C21 of the Spanish Ministry of Science and Innovation (MICINN).